\documentclass{article}
\usepackage[margin=1in]{geometry}

\usepackage{subcaption}  % Add subcaption package
\usepackage{amsmath}
\usepackage{dsfont}

\usepackage{graphicx} 
\usepackage[natbib=true, style=alphabetic]{biblatex}
\usepackage[utf8]{inputenc} % allow utf-8 input
\usepackage[T1]{fontenc}    % use 8-bit T1 fonts
\usepackage{hyperref}
\usepackage{amsmath}       % hyperlinks
\usepackage{url}            % simple URL typesetting
\usepackage{booktabs}       % professional-quality tables
\usepackage{amsfonts}       % blackboard math symbols
\usepackage{nicefrac}       % compact symbols for 1/2, etc.
\usepackage{microtype}      % microtypography
\usepackage{xcolor}         % colors
\usepackage{algorithm}
\usepackage{algpseudocode}
\newcommand{\norm}{\|}

\newcommand\abs[1]{\left|#1\right|}
\newcommand{\arrbegin}{\begin{eqnarray}}
\newcommand{\arrend}{\end{eqnarray}}

\newcommand{\del}{\partial}
\newcommand{\RR}{\mathbb{R}}

\newcommand{\NN}{\mathbb{N}}

\newcommand{\EE}{\mathop{\mathbb{E}}}
\newcommand{\PP}{\mathop{\mathbb{P}}}
\newcommand{\FF}{\mathbb{F}}

\newcommand{\holder}{\text{H{\"o}lder }}

\newcommand{\eps}{\epsilon}

\newcommand{\la}{\langle}
\newcommand{\ra}{\rangle}

\newcommand{\X}{\mathcal{X}}
\usepackage{bm}
\newcommand{\bx}{\bm{x}}
\newcommand{\by}{\bm{y}}
\newcommand{\bz}{\bm{z}}

\usepackage{mathtools}
\DeclarePairedDelimiter\ceil{\lceil}{\rceil}
\DeclarePairedDelimiter\floor{\lfloor}{\rfloor}

\bibliography{ref}

\title{Transfer Learning for Latent Variable Network Models}

\author{
  \begin{minipage}[t]{0.45\textwidth}
    \centering
    Akhil Jalan\thanks{Corresponding author.} \\
    Department of Computer Science \\
    UT Austin \\
    \texttt{akhiljalan@utexas.edu}
  \end{minipage}
  \hfill
  \begin{minipage}[t]{0.45\textwidth}
    \centering
    Arya Mazumdar \\
    Hal\i{}c{\i}o\u{g}lu Data Science Institute and Dept of CSE \\
    UC San Diego \\
    \texttt{arya@ucsd.edu}
  \end{minipage}
  \vspace{1em} \\ % Adjust space between rows
  \begin{minipage}[t]{0.45\textwidth}
    \centering
    Soumendu Sundar Mukherjee \\
    Statistics and Mathematics Unit (SMU) \\
    Indian Statistical Institute, Kolkata \\
    \texttt{ssmukherjee@isical.ac.in}
  \end{minipage}
  \hfill
  \begin{minipage}[t]{0.45\textwidth}
    \centering
    Purnamrita Sarkar \\
    Department of Statistics \& Data Sciences \\
    UT Austin \\
    \texttt{purna.sarkar@austin.utexas.edu}
  \end{minipage}
}

\usepackage{amsthm}
\usepackage{amsmath}
\usepackage{amssymb}

\newtheorem{theorem}{Theorem}[section]
\newtheorem{cor}[theorem]{Corollary}
\newtheorem{lemma}[theorem]{Lemma}
\newtheorem{prop}[theorem]{Proposition}
\newtheorem{defn}[theorem]{Definition}

\newtheorem{remark}[theorem]{Remark}
\newtheorem{assumption}[theorem]{Assumption}

\begin{document}

\maketitle

% !TEX root = ./neurips_2024.tex
\begin{abstract}
We study transfer learning for estimation in latent variable network models. In our setting, the conditional edge probability matrices given the latent variables are represented by $P$ for the source and $Q$ for the target. We wish to estimate $Q$ given two kinds of data: (1) edge data from a subgraph induced by an $o(1)$ fraction of the nodes of $Q$, and (2) edge data from all of $P$. If the source $P$ has no relation to the target $Q$, the estimation error must be $\Omega(1)$. However, we show that if the latent variables are shared, then vanishing error is possible. We give an efficient algorithm that utilizes the ordering of a suitably defined graph distance. Our algorithm achieves $o(1)$ error and does not assume a parametric form on the source or target networks. Next, for the specific case of Stochastic Block Models we prove a minimax lower bound and show that a simple algorithm achieves this rate. Finally, we empirically demonstrate our algorithm's use on real-world and simulated network estimation problems.
\end{abstract}

% We study transfer learning for graph reconstruction in a latent space model. In our setting, we wish to learn a target graph $Q$, given two kinds of data: (1) Edge data from within an $o(1)$ fraction of the $Q$ nodes, and (2) Edge data from all the nodes of a different graph $P$. If $P$ has no relation to $Q$, the normalized reconstruction error must be $\Omega(1)$. However, we show that if the latent variables are shared, then vanishing error is possible. In the case of stochastic block models, this condition simply states that the true clustering of $Q$ coarsens the clustering of $P$. We prove a minimax lower bound on transfer learning in this setting, and show that existing clustering algorithms can achieve the minimax rate efficiently. For a general latent variable model, a different approach is needed. We propose an efficient transfer learning algorithm that utilizes the ordering of a suitably defined graph distance. We give theoretical guarantees on our algorithm based on the distribution of latent positions, and empirically demonstrate its use on real-world and simulated graph transfer problems. 

% !TEX root = ./neurips_2024.tex
\section{Introduction}

Within machine learning and statistics, the paradigm of {\em transfer learning} describes a setup where data from a source distribution $P$ is  exploited to
improve estimation of a target distribution $Q$ for which a small amount of data is available. Transfer learning is quite well-studied in learning theory, starting with works such as \cite{ben2006analysis,cortes2008sample,crammer2008learning}, and at the same time has found applications in areas such as computer vision~\citep{tzeng2017adversarial} and speech recognition~\citep{huang2013cross}. 
A fairly large body of work in transfer learning considers different types of relations that may exist between $P$ and $Q$, for example,~\cite{mansour2009domain,hanneke2019value,hanneke2022no}, with emphasis on model selection, multitask learning and domain adaptation.
On the other hand, optimal nonparametric rates for transfer learning have very recently been studied, both for regression and classification problems~\citep{cai-wei-2021,cai-pu-2022}.

In this paper, we study transfer learning in the context of {\em  random network/graph models}. 
In our setting, we observe Bernoulli samples from the full $n \times n$ edge probability matrix for the source $P$ and only a $n_Q \times n_Q$ submatrix of $Q$ for $n_Q \ll n$. We would like to estimate the full $n \times n$ probability matrix $Q$, using the full source data and limited target data, i.e., we are interested in the task of estimating $Q$ in the partially observed target network, utilizing information from the fully observed source network. This is a natural extension of the transfer learning problem in classification/regression to a network context. However, it is to be noted that network transfer is a genuinely different problem owing to the presence of edge correlations. 

%\bl There is a vast body of empirical work on network transfer~\cite{Tang2016,cao2010transfer, Qiao2023SemisupervisedDA}. However, our setting, which is motivated by problems arising in Biology, not been studied in the literature. It is also different from the closest existing works~\cite{simchowitz-2023,Wu2018LinkPF}, where every node in the target network has some representative edges available; ~\cite{levin-lodhia-levina-2022} learn from multiple networks with the same expectation but different noise variances. In contrast, we consider a problem where we have a source network with the same latent variables. However, only a small induced subgraph sampled from the target link probability matrix is observed, and the edge probabilities can be different.  This comes up in a number of applications on biological networks, where data is scarce. Our main goal in this manuscript is to render a solid theoretical foundation to network transfer, by establishing minimax lower bounds as well as matching upper bounds achieved by efficient algorithms. (tone down a little? - Yes!) \bk We formally define this later.

%\rd [PS:]task of link prediction - runs the risk of people looking for link pred exps \bk

While transfer learning in graphs seems to be a fundamental enough problem to warrant attention by itself, we are also motivated by potential applications. For example, metabolic networks model the chemical interactions related to the release and utilization of energy within an organism \citep{Christensen2000}. Existing algorithms for metabolic network estimation~\citep{sen2018asapp,baranwal2020deep} and biological network estimation more broadly ~\citep{fan2019functional, li-2022-joint} typically assume that some edges are observed for every node in the target network. One exception is \cite{kshirsagar2013multitask}, who leverage side information for host-pathogen protein interaction networks. For the case of metabolic networks, determining interactions {\em in vivo}\footnote{In the organism, as opposed to {\em in vitro} (in the lab).} requires metabolite balancing and labeling experiments, so only the edges whose endpoints are {\em both} incident to the experimentally chosen metabolites are observed \citep{Christensen2000}. For a non-model organism, the experimentally tested metabolites may be a small fraction of all metabolites believed to affect metabolism. However, data for a larger set of metabolites might be available for a model organism. 
%, motivating transfer learning for biological networks \citep{kshirsagar2015combine}. \blue{Existing algorithms for metabolic network estimation~\citep{sen2018asapp,baranwal2020deep} and biological network estimation more broadly ~\citep{fan2019functional, li-2022-joint} typically require features and are not organism-specific.}

To study transfer learning on networks, one needs to fix a general enough class of networks that is appropriate for the applications (such as the biological networks mentioned above) and also suitable to capture the transfer phenomenon. The latent variable models defined below appear to be a natural candidate for that.

%\blue{latent positions also include SBM, dot product, MMSB, etc. Mention before Hoff}
{\bf Latent variable models.} Latent variable network models consist of a large class of models whose edge probabilities are governed by the latent positions of nodes. This includes latent distance models, stochastic block models, random dot product graphs and mixed membership block models~\citep{hoff_latent_2002,hoff2007modeling,handcock_model-based_2007,holland_stochastic_1983,rubin-dot-product-2022,airoldi08mixed}.
They can also be unified under graph limits or graphons~\citep{books/daglib/0031021,bickel2009nonparametric}, which provide a natural representation of vertex exchangeable graphs~\citep{aldous-representation-array,hoover-exchangeability}. For unseen latent variables $\bx_1, \dots, \bx_n \in \mathcal{X} \subset \RR^d$ and unknown function $f_Q: \mathcal{X} \times \mathcal{X} \to [0,1]$ where $\mathcal{X}$ is  a %closed bounded
compact set and $d$ an arbitrary fixed dimension, the edge probabilities are:
\begin{align}
Q_{ij} = f_Q(\bx_i, \bx_j).
\label{eq:latent-model}
\end{align}
%\EE[A_{ij}|\bx_i,\bx_j] = f(\bx_i, \bx_j) 

Typically, in network estimation, one observes adjacency matrix $\{A_{ij}\}$ distributed as $\{\textrm{Bernoulli}(Q_{ij})\}$, and either has to learn $\bx_i$ or directly estimate $f_Q$. There has been much work in the statistics community on estimating $\bx_i$ for specific models (usually up to rotation). For stochastic block models, see the excellent survey in~\cite{abbe2017community}. %\rd Cite mixed membership and RDPG embedding stuff\bk.  %In particular, let $g: \mathcal{X} \to \mathcal{X}$ be any invertible function. If $h(\bx, \by) := f(g(\bx), g(\by))$, then the data arising from function $f$ on latent positions $\bx_1, \dots, \bx_n$ cannot be distinguished from function $h$ on latent positions $g^{-1}(\bx_1), \dots, g^{-1}(\bx_n)$. 

Estimating $f_Q$ can be done with some additional assumptions~\citep{chatterjee2015matrix}. When $f_Q$ has appropriate smoothness properties, one can estimate it by a histogram approximation~\citep{olhede-wolfe-2014,chan2014consistent}. This setting has also been compared to nonparametric regression with an unknown design~\citep{gao2015rate}. Methods for network estimation include Universal Singular Value Thresholding~\citep{chatterjee2015matrix,xu2018rates}, combinatorial optimization~\citep{gao2015rate,klopp2017oracle}, and neighborhood smoothing~\citep{zhang-levina-zhu-2017,mukherjee2019graphon}.

{\bf Transfer learning on networks.} We wish to estimate the target network $Q$. However, we only observe $f_Q$ on $\binom{n_Q}{2}$ pairs of nodes, for a uniformly random subset of variables $S \subset \{1, 2, \dots, n\}$. We assume $S$ is vanishingly small, so $\abs{S} := n_Q = o(n)$.

Absent additional information, we cannot hope to achieve $o(1)$ mean-squared error. To see this, suppose $f_Q$ is a stochastic block model with $2$ communities of equal size. For a node $i \not \in S$, no edges incident to $i$ are observed, so its community cannot be learned. Since $n_Q \ll n$, we will attain $\Omega(1)$ error overall. To attain error $o(1)$, we hope to leverage transfer learning from a source $P$ if available. In fact, we give an efficient algorithm to achieve $o(1)$ error, formally stated in Section~\ref{sec:rankings}.
%Theorem~\ref{thm:alg1error}.
\begin{theorem}[Theorem~\ref{thm:alg1error}, Informal]
There exists an efficient algorithm such that, if given source data $A_P \in \{0,1\}^{n \times n}$ and target data $A_Q \in \{0,1\}^{n_Q \times n_Q}$ coming from an {\em appropriate} pair $(f_P, f_Q)$ of latent variable models, outputs $\hat Q \in \RR^{n \times n}$ such that: 
\[
\PP\bigg[\frac{1}{n^2} \norm Q - \hat Q \norm_F^2 \leq o(1)\bigg] \geq 1 - o(1).
\]
\label{thrm:main-informal}
\end{theorem}
There must be a relationship between P and Q for them to be an {\em appropriate} pair for transfer learning. We formalize this relationship below.

{\bf Relationship between source and target.} 
It is natural to consider pairs $(f_P, f_Q)$ such that for all $\bx, \by \in \mathcal{X}$, the difference $(f_P(\bx, \by) - f_Q(\bx, \by))$ is small. For example, \citet{cai-pu-2022} study transfer learning for nonparametric regression when $f_P - f_Q$ is close to a polynomial in $\bx, \by$.
But, requiring $f_P-f_Q$ to be pointwise small does not capture a broad class of pairs in the network setting. For example, if $f_P = \alpha f_Q$. Then $f_P - f_Q = (\alpha-1)f_Q$ can be far from all polynomials if $f_Q$ is, e.g. a \holder-smooth graphon.\footnote{In fact, \cite{cai-pu-2022} highlight this exact setting as a direction for future work.} However, under the network model, this means $A_P$ and $A_Q$ are stochastically identical modulo one being $\alpha$ times denser than the other. 

%So transfer should in fact be easy in this setting, i.e. if I know $P(A_P(i,j)=1|\bx_i,\bx_j)$, then corrsponding $Q$ probability is $1/\alpha$ times  the  former. 
We will therefore consider pairs $(f_P,f_Q)$ that are close in some measure of local graph structure. With this in mind, we use a graph distance  introduced in~\cite{mcs21} for a different inference problem.
\newcommand{\be}{\bm{e}}
\begin{defn}[Graph Distance]
Let $P \in [0,1]^{n \times n}$ be the probability matrix of a graph. For $i, j \in [n]$ we will use the the graph distance between $i, j$, $i\neq j$ as: 
\[
d_P(i, j) := \norm (\be_i - \be_j)^T P^2 (I - \be_i \be_i^T - \be_j \be_j^T)\norm_2^2.
\]
Where $\be_i, \be_j \in \RR^{n}$ are standard basis vectors. 
\label{defn:graph-distance}
\end{defn}
%\label{defn:graph-distance}

Intuitively, this first computes the matrix $P^2$ of common neighbors, and then computes the distance between two rows of the same (ignoring the diagonal elements). 
We will require that $f_P, f_Q$ satisfy a local similarity condition on the relative rankings of nodes with respect to this graph distance. Since we only estimate the probability matrix of $Q$, the condition is on the latent variables $\bx_1, \dots, \bx_n$ of interest.  The hope is that the proximity in graph distance reflects the proximity in latent positions.
\begin{defn}[Rankings Assumption at quantile $h_n$]
Let $(P, Q)$ be a pair of graphs evaluated on $n$ latent positions. We say $(P, Q)$ satisfy the rankings assumption at quantile $h_{n} \leq 1$ if there exists constant $C > 0$ such that
for all $j \neq i$, if $j$ belongs to the bottom $h_n$-quantile of $d_P(i, \cdot)$, then $j$ belongs to the bottom $Ch_n$-quantile of $d_Q(i, \cdot)$. 
\label{defn:ranks}
\end{defn}

Note that the condition is for relative, not absolute graph distances.

To illustrate Definition~\ref{defn:ranks}, consider stochastic block models $f_P, f_Q$ with $k_P \geq k_Q$ communities respectively. If nodes $i, j$ are in the same communities then $P \be_i = P \be_j$, so $d_P(i, j) = 0$. We require that $j$ minimizes $d_Q(i, \cdot)$. This occurs if and only if $d_Q(i, j) = 0$. Hence if $i, j$ belong to the same community in $P$, they are in the same community in $Q$. Note that the converse is not necessary; we could have $Q$ with $1$ community and $P$ with arbitrarily many communities. 

% With the relationship between 
the source and target defined by the rankings assumption, our contributions are as follows.

{\bf (1) Algorithm for Latent Variable Models.}
We provide an efficient Algorithm~\ref{alg:q-averaging-row-wise} for latent variable models with H{\"o}lder-smooth $f_P, f_Q$. The benefit of this algorithm is that it does not assume a  parametric form of $f_P$ and $f_Q$.  We prove a guarantee on its error in Theorem ~\ref{thm:alg1error}. 

{\bf (2) Minimax Rates.}  We prove a minimax lower bound for Stochastic Block Models (SBMs) in Theorem~\ref{thrm:minimax-sbm}. Moreover, we provide a simple Algorithm~\ref{alg:q-perfect-clustering} that attains the minimax rate for this class (Proposition~\ref{prop:q-perfect-clustering}). 

{\bf (3) Experimental Results on Real-World Data.} We test both of our algorithms on real-world metabolic networks and dynamic email networks, as well as synthetic data (Section~\ref{sec:experiments}).
% % (Section~\ref{sec:experiments}).
% \blue{todo sec exps ref}

All proofs are deferred to the Appendix.

%  % \input{proposed-work}

\subsection{Other Related work}

Transfer learning has recently drawn a lot of interest both in applied and theoretical communities. The notion of transferring knowledge from one domain with a lot of data to another with less available data has seen applications in epidemiology~\cite{Apostolopoulos2020Covid19AD}, computer vision~\cite{long2015fully,Tzeng2017AdversarialDD,huh2016makes,pmlr-v32-donahue14,transfer2020neyshabur}, natural language processing~\cite{Daum2007FrustratinglyED}, etc. For a comprehensive survey see~\cite{Zhuang2019ACS,Weiss2016ASO,kim2022transfer}.
Recently, there have also been advances in the theory of transfer learning~\cite{yang2013theory,transferTheory2020trip,pmlr-v195-agarwal23b,cai2021transfer,cai-pu-2024,systemtheorey2023tyler}.

In the context of networks, transfer learning is particularly useful since labeled data is typically hard to obtain.~\cite{Tang2016} develop an algorithmic framework to transfer knowledge obtained using available labeled connections from a source network to do link prediction in a target network.~\cite{lee2017transfer} proposes a deep learning framework for graph-structured data that incorporates transfer learning. They transfer geometric information from the source domain to enhance performance on related tasks in a target domain without the need for extensive new data or model training. The SGDA method~\cite{Qiao2023SemisupervisedDA} introduce adaptive shift parameters to mitigate domain shifts and propose pseudo-labeling of unlabeled nodes to alleviate label scarcity.~\cite{zou2021transfer} proposes to transfer features from the previous network to the next one in the dynamic community detection problem.~\cite{pmlr-v195-simchowitz23a} work on combinatorial distribution shift for matrix completion, where only some rows and columns are given. A similar setting is used for link prediction in egocentrically sampled networks in~\cite{Wu2018LinkPF}.~\cite{zhu2021transfer} train a graph neural network for transfer based on an ego-graph-based loss function. 
Learning from observations of the full network and additional information from a game played on the network~\cite{leng2020learning,rossi2022learning}. \cite{wu-2024-fixed} study graph transfer learning for node regression in the Gaussian process setting, where the source and target networks are fully observed. 

~\cite{levin2022recovering} proposes an inference method from multiple networks all with the same mean but different variances. While our work is related, we do not assume $\EE[P_{ij}] = \EE[Q_{ij}]$. \cite{cao2010transfer} do joint link prediction on a collection of networks with the same link function but different parameters. %\cite{leskovec2024inferring} studies proportional fitting based on observed marginals across time and the time-averaged network in a Poisson model of dynamic networks. \cite{wang2023dynamic} predict a target graph at time $t + 1$ given graphs and label information for sources and targets at times $1, 2, \dots, t$ using a dynamic Wasserstein distance. \rd but there are a vast literature on dynamic networks do we want to get into it?\bk

%\blue{Learning from observations of the full network and additional information from a game played on the network~\citep{leng2020learning,rossi2022learning}}.

%\blue{\cite{leskovec2024inferring} studies proportional fitting based on observed marginals across time, and the time-averaged network, in a Poisson model of dynamic networks. \cite{wang2023dynamic} predict a target graph at time $t + 1$ given graphs and label information for sources and targets at times $1, 2, \dots, t$ using a dynamic Wasserstein distance.} 

Another line of related but different work deals with multiplex networks~\citep{lee2014multiplex,lee2015towards,iacovacci2016extracting,cozzo2018multiplex} and dynamic networks~\cite{dynamicNIPS2005_ec8b57b0,dynamic_10.1214/18-SS121,dynamic_sewell2015latent,dynamic10.5555/3042573.3042815,leskovec2024inferring,wang2023dynamic}. 
One can think of transfer learning in clustering as clustering with side information. Prior works consider stochastic block models with noisy label information~\citep{mossel2016local} or oracle access to the latent structure~\citep{mazumdar-2017}. 
%By contrast, the side information is not another network, like ours. 
%studies classification in 2-community SBM with side information using a belief propagation algorithm.

%\blue{\citet{wu2024graph} study graph transfer learning in the Gaussian process framework where $f_P=f_Q$, but the distribution of $\mathcal{X}$ is different. However, they observe the full source and target networks and the goal is to do node regression.}

%\blue{}

%\blue{\cite{yang2013theory} part of Hanneke's line of work related to active learning} [this I cited]

{\bf Notation.} We use lowercase letters $a, b, c$ to denote (real) scalars, boldface $\bx, \by, \bz$ to denote vectors, and uppercase $A, B, C$ to denote matrices. Let $a \lor b := \max\{a, b\}$ and $a \land b := \min\{a, b\}$. For integer $n > 0$, let $[n] := \{1, 2, \dots, n\}$. For a subset $S \subset [n]$ and $A \in \RR^{n \times n}$, let $A[S,S] \in \RR^{\abs{S} \times \abs{S}}$ be the principal submatrix with row and column indices in $S$. We denote the $\ell_2$ vector norm as $\norm \bx \norm = \norm \bx \norm_2$, dot product as $\la \bx, \by \ra$, and Frobenius norm as $\norm A \norm = \norm A \norm_F$. For functions $f, g: \NN \to \RR$ we let $f \lesssim g$ denote $f = O(g)$ and $f \gtrsim g$ denote $f = \Omega(g)$. All asymptotics $O(\cdot), o(\cdot), \Omega(\cdot), \omega(\cdot)$ are with respect to $n_Q$ unless specified otherwise. 
%For reals $a, b$ 

% ...\blue{todo write out notations.}
 
% \input{algorithm-rank2}

% \input{proposed-work}

% !TEX root = ./neurips_2024.tex

\section{Estimating Latent Variable Models with Rankings}\label{sec:rankings}

In this section we present a computationally efficient transfer learning algorithm for latent variable models. Algorithm~\ref{alg:q-averaging-row-wise} learns the local structure of $P$ based on graph distances (Definition~\ref{defn:graph-distance}). For each node $i$ of $P$, it ranks the nodes in $S$ with respect to the graph distance $d_P(i, \cdot)$. For most nodes $i, j \in [n]$, none of the edges incident to $i$ or $j$ are observed in $Q$. Therefore, we estimate $\hat Q_{ij}$ by using the edge information about nodes $r, s \in S$ such that $d_P(i, r)$ and $d_P(j, s)$ are small. 

Formally, we consider a model as in Eq.~\eqref{eq:latent-model} with a compact latent space $\mathcal{X} \subset \RR^d$ and latent variables sampled i.i.d. from the normalized Lebesgue measure on $\X$. We set $\mathcal{X} = [0,1]^d$ without loss of generality and assume that functions $f: \X \times \X \to [0,1]$ are $\alpha$-H{\"o}lder-smooth.
\begin{defn}\label{defn:holder-rigorous}
Let $f: \X \times \X \to \RR$ and $\alpha > 0$. We say $f$ is $\alpha$-\holder smooth if there exists $C_\alpha > 0$ such that for all $\bx, \bx^\prime, \by \in \X$, 
\[
\sum\limits_{\kappa \in \NN^d: \sum_i \kappa_i = \floor*{\alpha}}
\abs{
\frac{\del^{\sum_i \kappa_i} f}{\del_{x_1}^{\kappa_1} \cdots \del_{x_d}^{\kappa_d}} (\bx, \by)
- \frac{\del^{\sum_i \kappa_i} f}{\del_{x_1}^{\kappa_1} \cdots \del_{x_d}^{\kappa_d}} (\bx^\prime, \by)
}
\leq C_\alpha \norm \bx - \bx^\prime \norm_2^{\alpha \land 1}
\]
\end{defn}
%  i.e., for all $\bx, \bx^\prime, \by \in \X$, 
% \[
% \abs{f(\bx, \by) - f(\bx^\prime, \by)} 
% \leq C_\alpha \norm \bx - \bx^\prime \norm^{\alpha \land 1},
% \]
% for $\alpha > 0$ and some constant $C_\alpha>0.$ 
% Technically this condition is necessary, but not sufficient to define $\alpha$-H{\"o}lder-smoothness for $\alpha > 1$. See Definition~\ref{defn:holder-rigorous} for a fully rigorous definition.
To exclude degenerate cases where a node may not have enough neighbors in latent space, we require the following assumption. \begin{assumption}[Assumption 3.2 of \cite{mcs21}]\label{assumption:mcs}
Let $G$ be a graph on $\bx_1, \dots, \bx_n$. There exist $c_2 > c_1 > 0$ and $\Delta_n = o(1)$ such that for all $\bx_i, \bx_j$, 
\[
c_1 \norm \bx_i - \bx_j \norm^{\alpha \land 1} - \Delta_n
\leq \frac{1}{n^3} d_G(i, j)
\leq c_2 \norm \bx_i - \bx_j \norm^{\alpha \land 1} 
\]
\end{assumption}
The second inequality follows directly from H{\"o}lder-smoothness, and the first is shown to hold for e.g. Generalized Random Dot Product Graphs, among others \citep{mcs21}.

% % and let $\mu_\X$ be the normalized Lebesgue measure without loss of generality.\footnote{The Heine-Borel theorem implies that any compact $\X$ is contained in a product of intervals $\prod\limits_{i = 1}^{d} [a_i, b_i]$ for some $a_i, b_i$. Moreover, Haar's theorem implies that the unique translation-invariant Borel measure on the compact abelian group $(\mathcal{X}, +)$ is simply the normalized Lebesgue measure.}
% %  Even though the nodes $i, j$ might both not belong to $S$ (meaning )
% %  Each of the nodes in $S$ are observed in $A_Q$, 

% %  For a fixed node $j \in S$, if $d_P(i, j)$ is small in a relative sense 

% % ranks the neighbors of a node $i \in [n]$ 

\begin{algorithm}[t]
\caption{$\hat Q$-Estimation for Latent Variable Models\label{alg:q-averaging-row-wise}}
\begin{algorithmic}[1]
    \State \textbf{Input:} $A_P \in \{0,1\}^{n \times n}, A_Q \in \{0,1\}^{n_Q \times n_Q}, S \subset [n]$ s.t. $\abs{S} = n_Q$.
    \State Initialize $\hat{Q} \in \RR^{n \times n}$ to be all zeroes.
    
    \State For all $i$, all $j \neq i$, compute graph distances: 
    \[
    d_{A_P}(i, j) := \norm (e_i - e_j)^T (A_P)^2 (I - e_i e_i^T - e_j e_j^T)\norm_2^2
    \]
    
    \State Fix a bandwidth $h \in (0,1)$ based on $n, n_Q$. 
    
    \For{$i = 1$ to $n$}
        \State Let $T_i^{A_P}(h) \subset S$ be bottom $h$-quantile of $S$ with respect to $d_{A_P}(i, \cdot)$.
        \If{$i \in S$}
            \State Update $T_i^{A_P}(h) \leftarrow T_i^{A_P}(h) \cup \{i\}$.
        \EndIf
    \EndFor
    
    \For{$i = 2$ to $n$}
        \For{$1 \leq j < i$}
            \State Compute $\hat Q_{ij} = \hat Q_{ji}$ by averaging: 
            \[
            \hat Q_{ij}:= \frac{1}{\abs{T_i^{A_P}(h)} \abs{T_j^{A_P}(h)}} \sum_{r \in T_i^{A_P}(h)} \sum_{s \in T_j^{A_P}(h)} A_{Q;rs}
            \]
        \EndFor
    \EndFor
    \State Return $\hat Q$.
    % \State \textbf{return} $\hat Q :=\Tilde Q + \Tilde Q^T$ 
\end{algorithmic}
\end{algorithm}

We establish the rate of estimation for Algorithm~\ref{alg:q-averaging-row-wise} below. 

\begin{theorem}
Let $\hat Q$ be as in Algorithm~\ref{alg:q-averaging-row-wise}. Let $\beta \geq \alpha > 0$ be the H{\"o}lder-smoothness parameters of $f_P, f_Q$ respectively and $c$ an absolute constant. Suppose $(P, Q)$ satisfy Definition~\ref{defn:ranks} at $h_n = c \sqrt{\frac{\log n_Q}{n_Q}}$ and $P$ satisfies Assumption~\ref{assumption:mcs} with $\Delta_n = O((\frac{\log n}{n_Q})^{\frac{1}{2} \lor \frac{\alpha \land 1}{d}})$. Then there exists absolute constant $C > 0$ such that:
\begin{align*}
\PP\bigg[\frac{1}{n^2} \norm \hat Q - Q \norm_F^2 \lesssim 
\bigg(\frac{d}{2}\bigg)^{\frac{\beta \land 1}{2}}\bigg(\frac{\log n}{n_Q}\bigg)^{\frac{\beta \land 1}{2d}}\bigg]
\geq 1 - n_Q^{-C}
\end{align*}
\label{thm:alg1error}
\end{theorem}
% \label{thm:erroralg1}
To parse Theorem~\ref{thm:alg1error}, consider the effect of various parameter choices. First, observe that our upper bound scales quite slowly with $n$. Even if $n$ is superpolynomial in $n_Q$, e.g. $n = n_Q^{\log n_Q}$, then $\log n = O((\log n_Q)^2) = n_Q^{o(1)}$, so the overall effect on the error is dominated by the $n_Q$ term. 

Second, the bound is worse in large dimensions, and scales exponentially in $\frac{1}{d}$. This kind of scaling also occurs in minimax lower bounds for nonparametric regression~\citep{Tsybakov:1315296}, and upper bounds for smooth graphon estimation~\citep{xu2018rates}. However, we caution that nonparametric regression can be quite different from network estimation; it would be very interesting to know the dependence of dimension on minimax lower bounds for network estimation, but to the best of our knowledge this is an open problem. Finally notice that a greater smoothness $\beta$ results in a smaller error, up to $\beta = 1$, exactly as in \citep{gao2015rate,klopp2017oracle,xu2018rates}.

% !TEX root = ./neurips_2024.tex

\section{Minimax Rates for Stochastic Block Models}

In this section, we will show matching lower and upper bounds for a very structured class of latent variable models, namely, Stochastic Block Models (SBMs). 
%SBMs can be modeled by Eq.~\ref{eq:latent-model} with $\X = [0, 1]$ and piecewise constant $f$. 

\begin{defn}[SBM]
Let $P \in [0,1]^{n \times n}$. We say $P$ is an $(n, k)$-SBM if there exist $B \in [0,1]^{k \times k}$ and $z: [n] \to [k]$ such that for all $i, j$, $P_{ij} = B_{z(i) z(j)}$. We refer to $z^{-1}(\{j\})$ as community $j \in [k]$. \label{defn:sbm-main}
\end{defn}
% \blue{AJ: define matrix version in appendix}
%and $Z$ has minimum column sum (community size) $n_{min}$. 
% for some $k$ depending on $n, n_{min}$,
% Moreover, let $a_0 = \min\limits_{i} B_{ii}$ and $a_1 = \max\limits_{i \neq j} B_{ij}$. Then $M$ has signal-to-noise ratio: $\frac{(a_0 - a_1)^2}{\sqrt{a_0(1-a_1)}} \geq s$. 

We first state a minimax lower bound, proved via Fano's method. 
\begin{theorem}[Minimax Lower Bound for SBMs]
Let $k_P \geq k_Q \geq 1$ with $k_Q$ dividing $k_P$. Let $\mathcal{F}$ be the family of pairs $(P, Q)$ where $P$ is an $(n, k_P)$-SBM, $Q$ is an $(n, k_Q)$-SBM, and $(P, Q)$ satisfy Definition~\ref{defn:ranks} at $h_n = 1/k_P$. Moreover, suppose $S \subset [n]$ is restricted to contain an equal number of nodes from communities $1, 2, \dots, k_P$ of $P$. Then the minimax rate of estimation is: 
\begin{align*}
\inf\limits_{\hat Q \in [0,1]^{n \times n}} \sup\limits_{(P, Q) \in \mathcal{F}} 
\EE\bigg[\frac{1}{n^2} \norm \hat Q - Q \norm_F^2 \bigg]
&\gtrsim \frac{k_Q^2}{n_Q^2}.
\end{align*}
\label{thrm:minimax-sbm}
\end{theorem}
Note that Definition~\ref{defn:ranks} at $h_n = 1/k_P$ implies that the true community structure of $Q$ coarsens that of $P$. The condition that $k_Q$ divides $k_P$ is merely a technical one that we assume for simplicity.

%We give a sketch of the proof below. 

% {\em Proof idea of Theorem~\ref{thrm:minimax-sbm}.} We use Fano's method. 
% We construct a family of a family of paired SBMs $(P_w, Q_w, S)$ indexed by words $w$ of a suitable error correcting code. The cluster structures of all $P_w$ and all $Q_w$ are identical, and moreover the clustering of $Q_w$ is a coarsening of the clustering of $P_w$ by Assumption~\ref{assumption:ranks}. However, even if we can perfectly cluster $Q_w$, the inter-community edge connections between $Q_w, Q_{w^\prime}$ for $w \neq w^\prime$ are far apart due to distance properties of the code. Moreover, learning inter-community edge connections within $P_w$ does not help learn those of $Q_w$. Therefore, estimation error is limited by the ability to learn $\binom{k_Q}{2}$ parameters with $\binom{n_Q}{2}$ observations. This can be bounded with Fano's Inequality. 

We remark that minimax lower bounds for smooth graphon estimation are established by first showing lower bounds for SBMs, and then constructing a graphon with the same block structure using smooth mollifiers \citep{gao2015rate}. Therefore, we expect that Theorem~\ref{thrm:minimax-sbm} can also be extended to the graphon setting, using the same techniques. However, sharp lower bounds for other classes such as Random Dot Product Graphs will likely require different techniques~\citep{xie2020optimal,yan2023minimax}. 

% we expect that our lower bound can be extended to the graphon regime using the techniques of \cite{gao2015rate}, and sparse graphons using \cite{klopp2017oracle}. 

% \begin{remark}[Comparison to nonparametric regime for ordinary SBM estimation]
% In \cite{gao2015rate}, the minimax rate for estimating an SBM on $n_Q$ nodes with $k_Q$ communities is shown to be $\frac{k_Q^2}{n_Q^2} + \frac{\log k_Q}{n_Q}$. Surprisingly, Theorem~\ref{thrm:minimax-sbm} shows that estimating $n^2$ entries of $Q$, rather than just the $n_Q^2$ observed entries, has the same minimax rate in the presence of additional information from $P$. In other words, while $P$ removes the bottleneck for clustering, the bottleneck for nonparametric regression remains. 
% \end{remark}

\begin{remark}[Clustering Regime]
In Appendix \ref{subsection:sbm-clustering-error} we also prove a minimax lower bound of $\frac{\log k_Q}{n_Q}$ in the regime where the error of recovering the true clustering $z$ dominates. This matches the rate of \cite{gao2015rate}, but for estimating all $n^2$ entries of $Q$, rather than just the $n_Q^2$ observed entries. 
\end{remark}
Theorem~\ref{thrm:minimax-sbm} suggests that a very simple algorithm might achieve the minimax rate. Namely, use both $A_P, A_Q$ to learn communities, and then use only $A_Q$ to learn inter-community edge probabilities. If $(P, Q)$ are in the nonparametric regime where regression error dominates clustering error (called the {\em weak consistency} or {\em almost exact recovery} regime), then the overall error will hopefully match the minimax rate. 

We formalize this approach in Algorithm~\ref{alg:q-perfect-clustering}, and prove that it does achieve the minimax error rate in the weak consistency regime. To this end, we define the signal-to-noise ratio of an SBM with parameter $B \in [0,1]^{k \times k}$ as follows:
\[
s := \frac{p-q}{\sqrt{p(1-q)}},
\]
where $p = \min_{i} B_{ii}, q = \max_{i \neq j} B_{ij}$.
\begin{algorithm}
\caption{$\hat Q$-Estimation for Stochastic Block Models}
\begin{algorithmic}[1]
    \State \textbf{Input:} $A_P \in \{0,1\}^{n \times n}, A_Q \in \{0,1\}^{n_Q \times n_Q}, S \subset [n]$ s.t. $\abs{S} = n_Q$.
    
    \State Estimate clusterings 
    $\hat Z_P \in \{0,1\}^{n \times k_P}, \hat Z_Q \in \{0,1\}^{n_Q \times k_Q}$ 
    using \cite{chen2014improved} on $A_P, A_Q$ respectively.

    \State Let $\hat W_Q \in \RR^{k_Q \times k_Q}$ be diagonal with 
    \[
    \hat W_{Q;ii} = (\bm{1}^T \hat Z_Q \bm{e}_i)^{-1}
    \]
    % \State Estimate clustering $$ using \cite{fei2019exponential} with $A_Q$.

    \State Initialize $\hat \Pi \in \{0,1\}^{k_P \times k_Q}$ to be all zeroes. 
    \For{$i \in S$}
        \State Let $j_P \in [k_P], j_Q \in [k_Q]$ be the unique column indices at which row $i$ of $\hat Z_P, \hat Z_Q$ respectively are nonzero. 
        \State Let $\hat \Pi_{j_P, j_Q} = 1$
    \EndFor

    % \State Estimate projection $\hat \Pi \in \{0,1\}^{k_P \times k_Q}$ using the Hungarian algorithm with loss function: 
    % \[\min\limits_{U \in \{0,1\}^{k_P \times k_Q}} \norm I_S Z_P U - I_S Z_Q \norm 
    % \]

    \State Let $\hat B_Q \in [0,1]^{k_Q \times k_Q}$ be the block-average: 
    \[
    \hat B_Q = \hat W_Q \hat Z_Q^T A_Q \hat Z_Q \hat W_Q 
    \]

    \State \textbf{return} $\hat Q := \hat Z_P \hat \Pi \hat B_Q \hat \Pi^T \hat Z_P^T$
\end{algorithmic}
\label{alg:q-perfect-clustering}
\end{algorithm}

\begin{prop}[Error rate of Algorithm~\ref{alg:q-perfect-clustering}]
Suppose $P, Q \in [0,1]^{n \times n}$ are $(n, k_P), (n, k_Q)$-SBMs with minimum community sizes $n_{\min}^{(P)}, n_{\min}^{(Q)}$ respectively. Suppose also that $(P, Q)$ satisfy Definition~\ref{defn:ranks} at $h_n = n_{\min}^{(P)} / n$. Then if the signal-to-noise ratios are such that: 
$s_P \geq C (\frac{\sqrt{n}}{n_{\min}^{(P)}} \lor \frac{\log^2(n)}{\sqrt{n_{\min}^{(P)}}})$
and $s_Q \geq C (\frac{\sqrt{n_Q}}{n_{\min^{(Q)}}} \lor \frac{\log^2(n_Q)}{\sqrt{n_{\min}^{(Q)}}})$
for large enough constant $C > 0$, Algorithm~\ref{alg:q-perfect-clustering} returns $\hat Q$ such that:
\[
\PP\bigg[\frac{1}{n^2} \norm 
\hat Q - Q
\norm_F^2  
\lesssim
\frac{k_Q^2 \log(n_{min}^{(Q)})}{n_Q^2}
\bigg] \geq 1 - O\bigg(\frac{1}{n_Q}\bigg).
\]
\label{prop:q-perfect-clustering}    
\end{prop}

\section{Experiments}\label{sec:experiments}
%\label{sec:experiments}

In this section, we test Algorithm~\ref{alg:q-averaging-row-wise} against several classes of simulated and real-world networks. We use quantile cutoff of $h_n = \sqrt{\frac{\log n_Q}{n_Q}}$ for Algorithm~\ref{alg:q-averaging-row-wise} in all experiments.

\begin{figure}[h!]
\centering
\includegraphics[width=\textwidth]{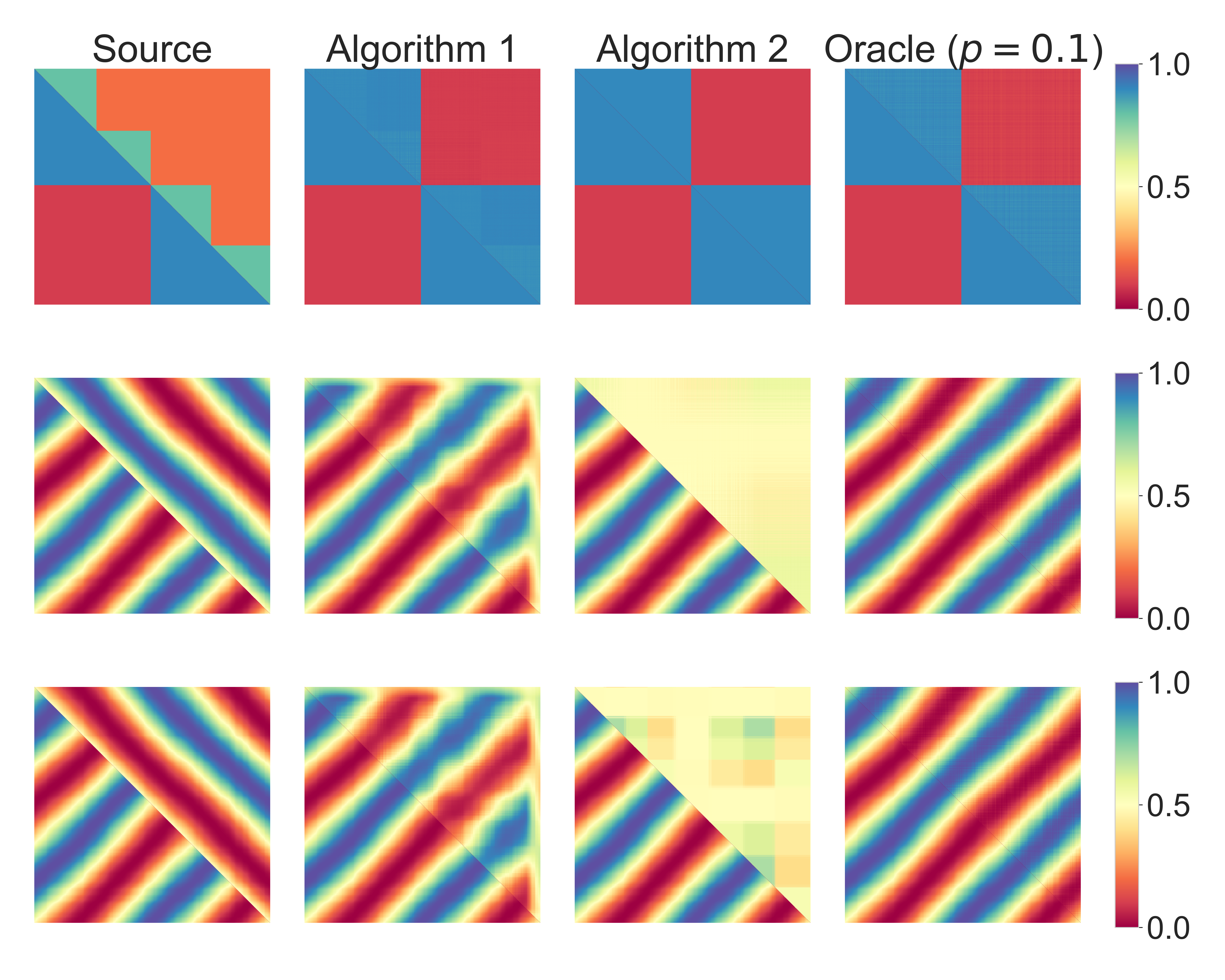}
\caption{Comparison of algorithms on three source-target pairs ($n = 2000, n_Q = 500$). In each heatmap, the lower triangle is the target $Q$. Algorithm~\ref{alg:q-perfect-clustering} performs best when $(P, Q)$ are SBMs (top), while Algorithm~\ref{alg:q-averaging-row-wise} is better for smooth graphons (2nd and 3rd row). 
\label{fig:heatmap}}
\end{figure}

% \subsection{Baseline Algorithms}
{\bf Baselines.} To the best of our knowledge, our exact transfer formulation has not been considered before in the literature. Therefore, we implement two algorithms as alternatives to Algorithm~\ref{alg:q-averaging-row-wise}. 
% \blue{AJ: say something about needing spectral and least squares to handle non-SBM inputs?}\\

{\em (1) Algorithm~\ref{alg:q-perfect-clustering}.} Given $A_P \in \{0,1\}^{n \times n}, A_Q \in \{0,1\}^{n_Q \times n_Q}$, let $k_P = \ceil*{\sqrt{n}}, k_Q = \ceil*{\sqrt{n_Q}}$. Compute spectral clusterings $\hat Z_P, \hat Z_Q$ with $k_P, k_Q$ clusters respectively. Let $J_S \in \{0,1\}^{n_Q \times n}$ is such that $J_{S;ij} = 1$ iff $i = j$ and $i \in S$. The projection $\hat \Pi \in \RR^{k_P \times k_Q}$ solves the least-squares problem $\min_{\Pi \in \RR^{k_P \times k_Q}} \norm J_S \hat Z_P \Pi - \hat Z_Q \norm_F^2$. We compute the $\hat{\Pi}$ differently from steps 4-7 in Algorithm~\ref{alg:q-perfect-clustering} to account for cases where $Q$ is not a true coarsening of $P$. When $Q$ is a true coarsening of $P$, this reduces to the procedure in steps 4-7. 

Given $\hat Z_P, \hat \Pi$ we return $\hat Q$ as in Algorithm~\ref{alg:q-perfect-clustering}.
% \blue{remark: we should chooose the correct kP, kQ here I guess for the algorithms.}

{\em (2) Oracle.} Suppose that an oracle can access data for $Q$ on {\em all} $n \gg n_Q$ nodes as follows. Fix an error probability $p_{\rm flip} \in (0,1)$. The oracle is given symmetric $A_Q^\prime \in \{0,1\}^{n \times n}$ with independent entries following a mixture distribution. For all $i, j \in [n]$ with $i < j$ let $X_{ij} \sim \textrm{Bernoulli}(p_{\rm flip})$ and $Y_{ij} \sim \textrm{Bernoulli}(Q(\bx_i, \bx_j))$. Then:
\[
A_{Q;ij}^\prime = \mathds{1}_{i \in S, j \in S} Y_{ij}
+ (1 - \mathds{1}_{i \in S, j \in S}) ((1 - X_{ij}) Y_{ij}
+ X_{ij} (1 - Y_{ij})).
\]
% \begin{align*}
% A_{Q;ij}^\prime = 
% \begin{cases}	
% \textrm{Bernoulli}(Q(\bx_i, \bx_j)) & i \in S, j \in S \\
% \textrm{Bernoulli}(Q(\bx_i, \bx_j)) & i \not\in S \text{ or } j \notin S \text{ with probability } 1 - p_{\rm flip} \\
% 1 - \textrm{Bernoulli}(Q(\bx_i, \bx_j)) & i \not\in S \text{ or } j \notin S \text{ with probability } p_{\rm flip}
% \end{cases}
% \end{align*}
% In other words, the data for $i, j \in S$ are unbiased, but the rest of the edges are flipped independently with probability $p_{\rm flip}$.

Given $A_Q^\prime$, the oracle returns the estimate from Universal Singular Value Thresholding on $A_Q^\prime$ \cite{chatterjee2015matrix}. As $p_{\rm flip} \to 0$, the error will approach $O(n^{\frac{-2\beta}{2 \beta + d}})$ for a $\beta$-smooth network on on $d$-dimensional latent variables~\citep{xu2018rates}, so the oracle will outperform any transfer algorithm.

\noindent{\bf Simulations.} We first test on several classes of simulated networks. For $n_Q = 50, n = 200$, we run $50$ independent trials for each setting. We report results for each setting in Table~\ref{table:simul-results}, and visualize estimates for stylized examples in Figure~\ref{fig:heatmap}.

{\em Smooth Graphons.} The latent space is $\X = [0,1]$. We consider graphons of the form $f_\gamma(x,y) = \frac{x^\gamma + y^\gamma}{2}$ where $P, Q$ have different $\gamma$. We denote this the $\gamma$-Smooth Graphon.

% \begin{table}
% \centering
% \begin{tabular}
% {|p{2cm}| p{2cm}| |p{1.4cm} |p{1.4cm} | p{1.4cm}| p{1.4cm} |p{1.4cm}|}
% \hline
% Source     
%     & Target     
%     & Alg.~\ref{alg:q-averaging-row-wise}
%     & Alg.~\ref{alg:q-perfect-clustering}
%     & Oracle ($p = 0.1$)
%     & Oracle ($p = 0.3$)
%     & Oracle ($p = 0.5$) \\
% \hline
% Noisy-MMSB $(0.7, 0.3, 0.01)$ & Noisy-MMSB $(0.9, 0.1, 0.01)$ &
% {\bf 0.7473$\pm$ 0.0648} & 
% $1.3761\pm 1.1586$ & 
% {\em 0.9556 $\pm$ 0.0633} &
% $2.2568 \pm 0.3107$ &
% $4.2212 \pm 0.2825$ \\
% \hline 
% $0.1$-Smooth Graphon & $0.5$-Smooth Graphon & 
% {\em 1.7656 $\pm$ 0.7494} &
% $4.5033 \pm 1.5613 $ &
% {\bf 0.5016 $\pm$ 0.0562} & 
% $2.4423 \pm 0.4574$ &
% $5.7774 \pm 0.7126$ \\ 
% \hline 
% $\RR^{10}$ Latent$(2.5)$ & $\RR^{10}$ Latent$(1.0)$ & 
% {\bf 0.5744 $\pm$ 0.1086} & 
% $1.1773 \pm 1.0481 $ &
% {\em 0.7715 $\pm$ 0.0456} & 
% $2.1822 \pm 0.2741$ &
% $4.3335 \pm 0.3476$ \\
% \hline
% \end{tabular}
% \caption{Comparison of different algorithms on simulated networks. Each cell reports $\hat \mu \pm 2 \hat \sigma$ of the mean-squared error over 50 independent trials. Error numbers are all scaled by $1e2$ for ease of reading. Bold: Best algorithm. Emphasis: Second-best algorithm.}
% \label{table:simul-results}
% \end{table}
% %\vspace{-1em}

{\em Mixed-Membership Stochastic Block Model.} Set $k_P = \floor*{\sqrt{n}}, k_Q = \floor*{\sqrt{n_Q}}$. The latent space $\mathcal{X}$ is the probability simplex $\mathcal{X} = \Delta_{k_P} := \{x \in [0,1]^{k_P}: \sum_i x_i = 1\} \subset \RR^{k_P}$. The latent variables $\bx_1, \dots, \bx_{n}$ are iid-Dirichlet distributed with equal weights $\frac{1}{k_P}, \dots, \frac{1}{k_P}$. Then $P_{ij} = \bx_i^T B_P \bx_j$ and $Q_{ij} = \Pi(\bx_i)^T B_Q \Pi(\bx_j)$, for connectivity matrices $B_P \in [0,1]^{k_P \times k_P}, B_Q \in [0,1]^{k_Q \times k_Q}$, and projection $\Pi: \Delta_{k_P} \to \Delta_{k_Q}$ for a fixed subset of $[k_P]$. For parameters $a, b, \eps \in [0,1]$ we generate $B \in [0,1]^{k \times k}$ by sampling $E \in \textrm{Uniform}(-\eps, \eps)^{k \times k}$ and set $B = \textrm{clip}((a-b) I + b \bm{1} \bm{1}^T + E, 0, 1)$. We call this Noisy-MMSB$(a, b, \eps)$. 

{\em Latent Distance Model.} The latent space is the unit sphere $\mathcal{X} = \mathbb{S}^{d-1} \subset \RR^{d}$. For scale parameter $s > 0$, we call $f_s(\bm{x}, \bm{y}) = \exp(-s \norm \bx - \by \norm_2)$ the $\RR^{d}$-Latent$(s)$ model.

\begin{table}[t]
\centering
\setlength{\arrayrulewidth}{0.1pt} % Adjust the line thickness here
\begin{tabular}{p{2cm} p{2cm} p{1.4cm} p{1.4cm} p{1.4cm} p{1.4cm} p{1.4cm}}
\toprule
Source     
    & Target     
    & Alg.~\ref{alg:q-averaging-row-wise}
    & Alg.~\ref{alg:q-perfect-clustering}
    & Oracle ($p = 0.1$)
    & Oracle ($p = 0.3$)
    & Oracle ($p = 0.5$) \\
\midrule
Noisy-MMSB $(0.7, 0.3, 0.01)$ & Noisy-MMSB $(0.9, 0.1, 0.01)$ &
{\bf 0.7473$\pm$ 0.0648} & 
$1.3761\pm 1.1586$ & 
{\em 0.9556 $\pm$ 0.0633} &
$2.2568 \pm 0.3107$ &
$4.2212 \pm 0.2825$ 
\vspace{0.3cm} \\ 
$0.1$-Smooth Graphon & $0.5$-Smooth Graphon & 
{\em 1.7656 $\pm$ 0.7494} &
$4.5033 \pm 1.5613 $ &
{\bf 0.5016 $\pm$ 0.0562} & 
$2.4423 \pm 0.4574$ &
$5.7774 \pm 0.7126$ 
\vspace{0.3cm} \\ 
$\RR^{10}$ Latent$(2.5)$ & $\RR^{10}$ Latent$(1.0)$ & 
{\bf 0.5744 $\pm$ 0.1086} & 
$1.1773 \pm 1.0481 $ &
{\em 0.7715 $\pm$ 0.0456} & 
$2.1822 \pm 0.2741$ &
$4.3335 \pm 0.3476$ \\
\bottomrule
\end{tabular}
\vspace{0.3cm} 
\caption{Comparison of different algorithms on simulated networks. Each cell reports $\hat \mu \pm 2 \hat \sigma$ of the mean-squared error over 50 independent trials. Error numbers are all scaled by $1e2$ for ease of reading. Bold: Best algorithm. Emphasis: Second-best algorithm.}
\label{table:simul-results}
\end{table}

% Estimating metabolic network of iJN1463 ({\em Pseudomonas putida}) different different sources. Left: Source iWFL1372 ({\em Escherichia coli W}). 
% Right: Source iPC815 ({\em Yersinia pestis}). Shaded regions denote $[1, 99]$ percentile outcomes from $50$ trials.

{\bf Discussion.} When the latent dimension is larger than $1$ (the Noisy MMSB and Latent Variable Models), our Algorithm~\ref{alg:q-averaging-row-wise} is better than both Algorithm~\ref{alg:q-perfect-clustering} and the Oracle with $p = 0.1$. Note that Algorithms~\ref{alg:q-averaging-row-wise} and ~\ref{alg:q-perfect-clustering} use
$\frac{n_Q^2}{n^2} \approx 0.06$ unbiased edge observations from $Q$, while the Oracle with $p = 0.1$ observes $(1 - p) \frac{n^2 - n_Q^2}{n^2} \approx 0.9$ unbiased edge observations in expectation.

% Inverted Periodic Graphon & Periodic Graphon & 
% $0.121879 \pm 0.015151$ &  
% $0.141990 \pm 0.032075 $ &
% $0.009292 \pm 0.000550$ & 
% $0.117140 \pm 0.002813$ &
% $0.128060 \pm 0.004063$ \\ 
% \hline \\
% \blue{Maybe remove the liza row.}
% \blue{todo add bolds and caption to table.}

% \blue{todo insert heatmaps}

% \blue{Todo add col. labels to heatmaps.}

\begin{figure}[h!]
\centering
\includegraphics[width=\textwidth]{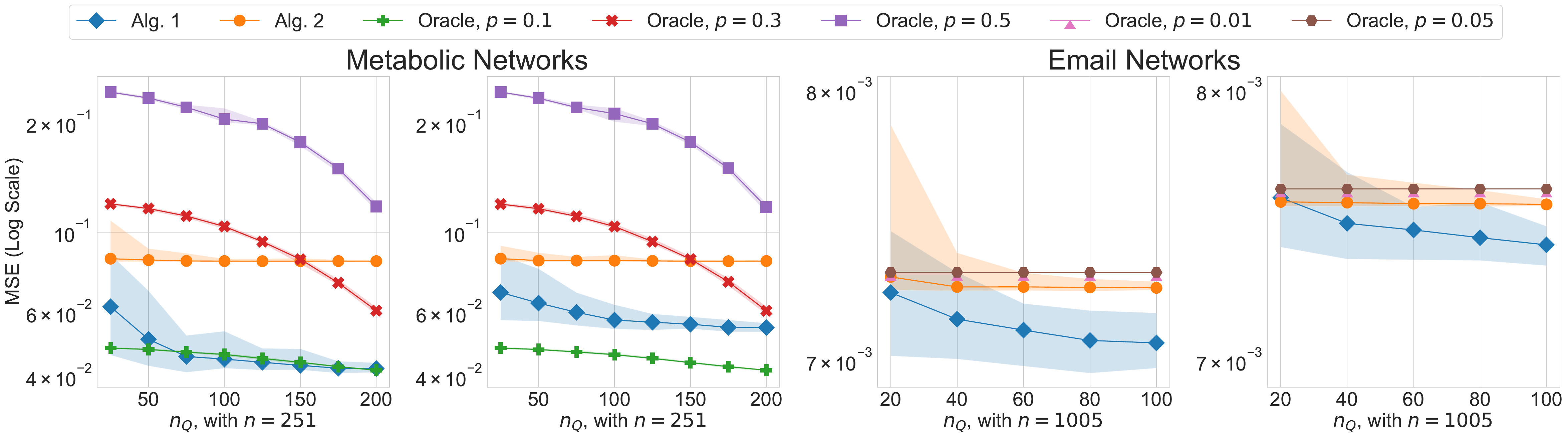}
\caption{Results of network estimation on real-world data. Shaded regions denote $[1, 99]$ percentile outcomes from $50$ trials. \\
{\em Left half}: Estimating metabolic network of iJN1463 ({\em Pseudomonas putida}) with source iWFL1372 ({\em Escherichia coli W}) leftmost, and source iPC815 ({\em Yersinia pestis}) second-left. \\
{\em Right half}: Using source data from days $1-80$ of \textsc{Email-EU} to estimate target days $81-160$ (third-left) and target days $561-640$ (rightmost). 
Note that we use smaller values of $p_{flip}$ for the Oracle in \textsc{Email-EU}.}
\label{fig:realworld}
\end{figure}

{\bf Real-World Data.} Next, we test on two classes of real-world networks. We summarize our dataset characteristics in Table~\ref{table:dataset-characteristics}. See Appendix~\ref{appendix:experiment-details} for further details.

\begin{table}[h!]
\centering
\caption{Dataset Characteristics}
\begin{tabular}{l c c l}
\toprule
Name & $n$ & Median Degree & Type \\
\midrule
BiGG Model iWFL1372 & 251 & 15.00 & Source  \\
BiGG Model iPC815 & 251 & 12.00 & Source  \\
BiGG Model iJN1463 & 251 & 14.00 & Target  \\
\textsc{Email-EU} Days 1-80 & 1005 & 6.92 & Source  \\
\textsc{Email-EU} Days 81-160 & 1005 & 7.35 & Target  \\
\textsc{Email-EU} Days 561-640 & 1005 & 7.66 & Target  \\
\bottomrule
\end{tabular}
\label{table:dataset-characteristics}
\end{table}

% \begin{table}[h!]
% \centering
% \caption{Dataset Characteristics}
% \begin{tabular}{l c c l}
% \toprule
% Name & $n$ & Median Degree & Type \\
% \midrule
% BiGG Model iWFL1372 & $251$ & $15.00$ & Source  \\
% BiGG Model iPC815 & $251$ & $12.00$ & Source  \\
% BiGG Model iJN1463 & $251$ & $14.00$ & Target  \\
% \textsc{Email-EU} Days 1-80 & $1005$ & $6.92$ & Source  \\
% \textsc{Email-EU} Days 81-160 & $1005$ & $7.35$ & Target  \\
% \textsc{Email-EU} Days 561-640 & $1005$ & $7.66$ & Target  
% \bottomrule
% \end{tabular}
% \label{table:dataset-characteristics}
% \end{table}

% \subsection{Real-World Data}

{\bf Transfer Across Species in Metabolic Networks.} For a fixed organism, a metabolic network has a node for each metabolite, and an edge exists if and only if two metabolites co-occur in a metabolic reaction in that organism. We obtain the unweighted metabolic networks for multiple gram-negative bacteria from the BiGG genome-scale metabolic model dataset \citep{bigg-models,norsigian2020bigg}. In the left half of Figure~\ref{fig:realworld}, we compare two choices of source organism in estimating the network for BiGG model iJN1463 ({\em Pseudomonas putida}). For a good choice of source, Algorithm~\ref{alg:q-averaging-row-wise} is competitive with the Oracle at $p=0.1$. 
%All metabolites in our data share the same Systems Biology Ontology. One useful feature we do have is the spatial compartment of a reaction (e.g. endoplasmic reticulum, mitochondria); however, we are interested in learning the entire network, and a single metabolite can participate in multiple compartments. 
% \blue{todo check this and provide an example in our data.} 
% We also note that our method only requires metabolic network data for metabolites of interest. Unlike other works, we do not explicitly learn latent space representations \cite{fan2019functional} or joint embeddings \cite{li-2022-joint} of the networks. 

 % This motivates our algorithm, which can estimate an unknown target organism by using data from a source (model) organism.

%Results are in Figures~\ref{fig:cho-with-baseline}. 

% \blue{Include an appendix table with species codes in BiGG, name, description.}
% \begin{figure}[h!]
% \centering
% \includegraphics[width=0.8\textwidth]{figs/metabolic_iWFL_1372_left_iPC815_right_target_iJN1463.pdf}
% \caption{Estimating metabolic network of iJN1463 ({\em Pseudomonas putida}) different different sources. Left: Source iWFL1372 ({\em Escherichia coli W}). 
% Right: Source iPC815 ({\em Yersinia pestis}). Shaded regions denote $[1, 99]$ percentile outcomes from $50$ trials.}
% \label{fig:gram-negative-with-baseline}
% \end{figure}
% \blue{todo -  give justification for the different sources and targets transferring differently. cite bio papers.}
% \begin{figure}
% \centering
% \end{figure}
{\bf Transfer Across Time in the Email Interaction Networks.} We use the \textsc{Email-EU} interaction network between $n = 1005$ members of a European research institution across $803$ days \cite{snapnets, motifs-2017}. The source graph $A_P$ is the network from day $1$ to $\approx 80$ ($[1,80]$). In Figure~\ref{fig:realworld} we simulate transfer with targets $[81, 160]$ (left) and $[561, 640]$ (right). We visualize results for arbitrary target periods; similar results hold for other targets.
%gives the results
% The data are timestamped across $803$ days; we bin the data into $10$ time periods $[t_{0}, t_{1}], \dots, [t_{9}, t_{10}]$ based on equally spaced timestamp percentiles. At a fixed time interval, there is an edge $\{u, v\}$ if and only if $u$ and $v$ interacted in that period. Our goal is to estimate the full interaction network at interval $[t_{\delta}, t_{\delta + 1}]$ using the full interaction newtork at $[t_{0}, t_{1}]$ and the subgraph at $n_Q << n$ nodes at $[t_{\delta}, t_{\delta + 1}]$, for $\delta > 0$. Figure~\ref{fig:email-eu} gives the results. 
Unlike metabolic networks, Algorithm~\ref{alg:q-perfect-clustering} has comparable performance to both our Algorithm~\ref{alg:q-averaging-row-wise} and the oracle algorithm with $p \in \{0.01, 0.05\}$. Compared to the metabolic networks, this indicates that the email interaction networks are relatively well-approximated by SBMs, although Algorithm~\ref{alg:q-averaging-row-wise} is still the best.
% \begin{figure}[h!]
% \centering
% \includegraphics[width=0.8\textwidth]{figs/email_source_t0_targets_t1_t7_smaller_p.pdf}
% \caption{Estimating a $1005$-node email interaction network with source data from day $1$ to $80$, and target from days $81$ to $160$ (left) and $561$ to $640$ (right). Shaded regions denote $[1, 99]$ percentile outcomes from $50$ trials.}
% \label{fig:email-eu}
% \end{figure}

% \blue{Todo fix x axis cutoffs.} 

% \begin{table}
%   \caption{Sample table title}
%   \label{sample-table}
%   \centering
%   \begin{tabular}{lll}
%     \toprule
%     \multicolumn{2}{c}{Part}                   \\
%     \cmidrule(r){1-2}
%     Name     & Description     & Size ($\mu$m) \\
%     \midrule
%     Dendrite & Input terminal  & $\sim$100     \\
%     Axon     & Output terminal & $\sim$10      \\
%     Soma     & Cell body       & up to $10^6$  \\
%     \bottomrule
%   \end{tabular}
% \end{table}

% !TEX root = ./neurips_2024.tex
\vspace{-1em}
\section{Conclusion}

In this paper, we study transfer learning for network estimation in latent variable models. We show that there exists an efficient Algorithm~\ref{alg:q-averaging-row-wise} that achieves vanishing error even when $n \geq n_Q^{\omega(1)}$, and a simpler Algorithm~\ref{alg:q-perfect-clustering} for SBMs that achieves the minimax rate. 

There are several interesting directions for future work. 
First, we believe that Algorithm~\ref{alg:q-averaging-row-wise} works for moderately sparse networks with population edge density $O(\frac{1}{\sqrt{n}})$. It would be interesting to see if a similar approach can work for edge density $O(\frac{\log n}{n})$, using a suitably modified graph distance~\citep{mcs21}. 
Second, the case of multiple sources is also interesting. We have focused on the case of one source distribution, as in \cite{cai2021transfer,cai-pu-2024}, but expect that our algorithms can be extended to multiple sources as long as they satisfy Definition~\ref{defn:ranks}. 

% Some directions for future work include handling sparse observations, and handling multiple source distributions. 

%For clarity we focus on the case of one source distribution, as in \cite{cai2021transfer,cai-pu-2022}. As in those works, extending our results to multiple source distributions $P_1, \dots, P_n$ should be straightforward. 

% Minimax lower bound for $d$ dimensions is open.

% \input{acknowledgments}

% \section*{References}

% plainnat doesn't work 
% plain doesn't work 
% abbrv and alpha work 
% \bibliographystyle{alpha}

%%%%%%%%%%%%%%%%%%%%%%%%%%%%%%%%%%%%%%%%%%%%%%%%%%%%%%%%%%%%

\section{Acknowledgments}

We gratefully acknowledge NSF grants 2217069, 2019844, and DMS 2109155.

AM was supported by NSF awards 2217058 and 2133484.

SSM was partially supported by an INSPIRE research grant (DST/INSPIRE/04/2018/002193) from the Dept. of Science and Technology, Govt. of India
and a Start-Up Grant from Indian Statistical Institute.

\printbibliography

\appendix
% !TEX root = ./neurips_2024.tex
\renewcommand{\bx}{\bm{x}}
\renewcommand{\by}{\bm{y}}

\section{Proofs}\label{appendix:proofs}

\subsection{Preliminaries}

Recall Hoeffding's inequality. 
\begin{lemma}[\cite{hoeffding1994probability}]
Let $X_1, \dots, X_n$ be independent random variables such that $a_i \leq X_i \leq b_i$ almost surely for all $i \in [n]$. Then: 
\[
\PP[\abs{
\sum\limits_{i = 1}^{n} (X_i - \EE[X_i])
} \geq t] \leq 2 \exp
\bigg(\frac{-2t^2}{\sum\limits_{i=1}^{n} (b_i - a_i)^2}
\bigg)
\]
\end{lemma}

We also need Bernstein's inequality. 
\begin{lemma}[Bernstein's Inequality]
Let $X_1, \dots, X_n$ be independent mean-zero random variables with $\abs{X_i} \leq 1$ for all $i$ and $n \geq 5$. Then: 
\[
\PP[\abs{
\frac{1}{n} \sum\limits_{i = 1}^{n} X_i
} \geq t] \leq 2 \exp
\bigg(\frac{- n t^2}{2(1 + \frac{t}{3})}
\bigg)
\leq 2 \exp(-\frac{n t^2}{4})
\]
\end{lemma}

\subsection{Proof of Theorem~\ref{thm:alg1error}}\label{appendix:rankings-algo}
% \blue{todo. (i) account for $\Delta_n$. (ii) write out the beginning of the bernoulli and smoothing error thing. (iii) clean up the various Bernstein and Hoeffdings, just go ahead and write them out clearly.}

Throughout this section, let $\X = [0, 1]^d$ and $\mu: \X \to [0,1]$ be the normalized Lebesgue measure.
% {\bf Different distance.} We will use a modification of Algorithm~\ref{alg:q-averaging-row-wise} with different notion of graph distance. If $i, j$ are nodes in $[n]$ then the population distance is: 
% \[
% d_P(i, j) := \sum\limits_{k \in [n]: k \neq i, j} ((P^2)_{ik} - (P^2)_{jk})^2
% = \norm (e_i - e_j)^T P^2 (I - e_i e_i^T - e_j e_j^T)\norm_2^2

% The formal definition of a \holder-smooth function is as follows.
% \begin{defn}\label{defn:holder-rigorous}
% Let $f: \X \times \X \to \RR$ and $\alpha > 0$. We say $f$ is $\alpha$-\holder smooth if there exists $C_\alpha > 0$ such that for all $\bx, \bx^\prime, \by \in \X$, 
% \[
% \sum\limits_{\kappa \in \NN^d: \sum_i \kappa_i = \floor*{\alpha}}
% \abs{
% \frac{\del^{\sum_i \kappa_i} f}{\del_{x_1}^{\kappa_1} \cdots \del_{x_d}^{\kappa_d}} (\bx, \by)
% - \frac{\del^{\sum_i \kappa_i} f}{\del_{x_1}^{\kappa_1} \cdots \del_{x_d}^{\kappa_d}} (\bx^\prime, \by)
% }
% \leq C_\alpha \norm \bx - \bx^\prime \norm_2^{\alpha \land 1}
% \]
% \end{defn}

We require the following Lemmata.
\begin{lemma}
Let $\upsilon \in (0, 1)$ and $\mu: \X \to [0,1]$ be the normalized Lebesgue measure. Then for all $\bx \in \X$, 
\[
\mu(\mathrm{Ball}(\bx, 2 \upsilon) \cap \mathcal{X}) 
\geq \mu(\mathrm{Ball}(\bm{0}, \upsilon) \cap \mathcal{X})
\] 
\label{lemma:d-dim-ball}
\end{lemma}

\begin{proof}
Recall $\X = [0, 1]^d$. Fix $\bx \in \X, \upsilon > 0$. 
Note that $\mu(\mathrm{Ball}(\bx, \upsilon) \cap \mathcal{X})$ is smallest when $\bx$ is a vertex of the hypercube; therefore take $\bx \in \{0,1\}^d$ without loss of generality. 
Then, note that for each $\bz \in \mathrm{Ball}(\bx, \upsilon) \cap \mathcal{X}$, we can find $(2^d -1)$ other points $\bz^\prime \in \mathrm{Ball}(\bx, \upsilon) \setminus \mathcal{X}$ by reflecting subsets of coordinates of $\bz$ about $\bx$. There are $2^d -1$ such nonempty subsets of coordinates. 
This shows that $\mu(\mathrm{Ball}(\bx, \upsilon) \cap \mathcal{X}) \geq \mu(\mathrm{Ball}(\bx, \upsilon))/2^d$ 
for all $\bx$. Since $\mu(\mathrm{Ball}(\bx, \upsilon)) \asymp \upsilon^d$, the conclusion follows. 
\end{proof}
% The volume of $\mu(\mathrm{Ball}(\bx, \upsilon))$ is exactly $\frac{\pi^{d/2}}{\Gamma(d/2 + 1)} \upsilon^d \asymp (\upsilon \sqrt{\frac{c}{d}})^{d}$ for a constant $c$ not depending on $d$ \citep{vershynin2018high}. Hence $\mu(\mathrm{Ball}(\bx, \upsilon) \cap \mathcal{X}) \geq (\frac{\nu \sqrt{c}}{2\sqrt{d}})^d$. The conclusion follows. 
% \blue{todo finish...what does this sign point mean exactly? should add a factor of two indeed.}

% \blue{AM: I believe this is a minor issue that can be easily  fixed. First note that, $\mu(\mathrm{Ball}(\bx, \upsilon) \cap \mathcal{X})$ is smallest when $\bx$ is a vertex of the hypercube; let us take that to be the origin without loss of generality. This shows that $\mu(\mathrm{Ball}(\bx, \upsilon) \cap \mathcal{X}) \ge \mu(\mathrm{Ball}(\bx, \upsilon))/2^d$ for all $\bx$. This will just add a factor of $2$ in lemma C.8.}

We will repeatedly make use of the concentration of latent positions. 
\begin{lemma}[Latent Concentration]
Let $\mathcal{X} = [0,1]^d$ and $\mu$ denote the normalized Lebesgue measure on $\mathcal{X}$. Suppose $\bx_1, \dots, \bx_n \sim \mathcal{X}$ are sampled iid and uniformly at random from $\mu$. Fix some $T \subset \mathcal{X}$ such that $\mu(T) = v$. Then, 
\begin{align*}
\PP[
\abs{vn - \abs{\{j \in [n]: \bx_j \in T\}}}
\geq 10 \sqrt{\frac{\log n}{n}}
] \leq n^{-10}
\end{align*}
\label{lemma:latent-conc-d-dim}
\end{lemma}
\begin{proof}
Let $X_i$ be an indicator variable that equals $1$ if $\bx_i \in T$ and zero otherwise. Notice the $X_i$ are iid and bounded within $[0,1]$. Moreover, $\sum_i \EE[X_i] = n \mu(T)$. Therefore by Hoeffding's inequality for any $t > 0$, 
\begin{align*}
\PP[
\abs{vn - \abs{\{j \in [n]: \bx_j \in T\}}}
\geq t
] \leq 2 \exp(\frac{-2t^2}{n})
\end{align*}
Setting $t = 10 \sqrt{\frac{\log n}{n}}$ gives the result. 	
\end{proof}

% \blue{Some subtlety about the corners of $[0,1]^d$.
% }

% \blue{todo simplify this one.}
\begin{cor}
Let $\eps > 0$. For $i \in [n]$ let $\eps_i^\prime > 0$ be $\eps_i^\prime := \sup\{\upsilon > 0: \mu(\mathrm{Ball}(\bx_i, \upsilon) \cap \mathcal{X}) \leq \eps$. 
Let $T_i := \mathrm{Ball}(\bx_i, \eps_i^\prime) \cap \mathcal{X}$. Let $u_i(S) := \abs{\{j \in S: \bx_j \in T_i\}}$ denote the number of members of $S$ landing in $T_i$. We claim that: 
\begin{align*}
\PP[\forall i \in [n]: \abs{u_i(S) - n_Q \eps} \leq 10 \sqrt{\frac{\log n}{n_Q}}] \geq 1 - n^{-8}	
\end{align*}
\label{cor:latent-conc-S-epsilon-ball}
\end{cor}
\begin{proof}
Notice that each $T_i$ has Lebesgue measure $\eps$ by definition. Therefore $\EE[u_i(S)] = n_Q \eps$. Since $S$ has $n_Q$ members, setting $t = 10 \sqrt{\frac{\log n}{n_Q}}$ in the statement of Lemma~\ref{lemma:latent-conc-d-dim} and taking a union bound over all $i \in [n]$ gives the conclusion.
\end{proof}

% From Lemma~\ref{lemma:d-dim-ball} we obtain the following. 

% \begin{cor}
% Let $\eps > 0$. For $i \in [n]$ let $\eps_i^\prime > 0$ be $\eps_i^\prime := \sup\{\upsilon > 0: \mu(\mathrm{Ball}(\bx_i, \upsilon) \cap \mathcal{X}) \leq \eps$. 
% Let $T_i := \mathrm{Ball}(\bx_i, \eps_i^\prime) \cap \mathcal{X}$. Let $u_i(S) := \abs{\{j \in S: \bx_j \in T_i\}}$ denote the number of members of $S$ landing in $T_i$. We claim that: 
% \begin{align*}
% \PP[\forall i: \abs{u_i(S) - n_Q \eps} \leq 10 \sqrt{\frac{\log n}{n_Q}}] \geq 1 - n_Q^{-8}	
% \end{align*}
% \label{cor:latent-conc-S-epsilon-ball}
% \end{cor}
% \begin{proof}
% Notice that each $T_i$ has Lebesgue measure $\eps$ by definition. Therefore $\EE[u_i(S)] = n_Q \eps$. Since $S$ has $n_Q$ members, setting $t = 10 \sqrt{\frac{\log n}{n}}$ in the statement of Lemma~\ref{lemma:latent-conc-d-dim} and taking a union bound over all $i \in [n]$ gives the conclusion.
% \end{proof}

We will decompose the error of Algorithm~\ref{alg:q-averaging-row-wise} into two parts. 
\begin{prop}
Let $\hat Q \in [0,1]^{n \times n}$ be the estimator from Algorithm~\ref{alg:q-averaging-row-wise}. Then: 
\[
\frac{1}{n^2} \norm Q - \hat Q \norm_F^2 
\leq \frac{2}{n^2} \sum\limits_{i, j \in [n]} (J_S(i,j) + J_B(i,j))
\]
Where $J_S, J_B$ are the smoothing and Bernoulli errors respectively: 
\begin{align*}
J_S(i,j) &:= \frac{1}{\abs{T_i}^2 \abs{T_j}^2}
\bigg(\sum\limits_{r \in T_i, s \in T_j} Q_{ij} - Q_{rs}
\bigg)^2 
\\
J_B(i,j) &:= \frac{1}{\abs{T_i}^2 \abs{T_j}^2} 
\bigg(\sum\limits_{r \in T_i, s \in T_j} Q_{rs} - A_{Q;rs}
\bigg)^2
\end{align*}
\label{prop:error-decomp}
\end{prop}

Controlling the Bernoulli errors is relatively straightforward.

\begin{prop}
Let $h$ be the bandwidth of Algorithm~\ref{alg:q-averaging-row-wise}. The Bernoulli error is at most $O(\frac{\log n}{m})$ with probability $\geq 1 - n^{-8}$, where $m = h^2 n_Q^2$. 
\end{prop}
\begin{proof}
Fix $i, j \in [n]$. We will bound the maximum Bernoulli error $J_S(i,j)$ over $i, j$, which suffices to bound the average. Let $m = \abs{T_i}\abs{T_j}$. We want to bound: 
\[
\abs{\frac{1}{\abs{T_i}\abs{T_j}} \sum\limits_{r \in T_i, s \in T_j}
(Q_{rs} - A_{Q;rs})}^2
\]
Notice each summand is bounded within $\pm \frac 1 m$. Bernstein's inequality gives: 
\[
\PP\bigg[\bigg(\frac{1}{\abs{T_i}\abs{T_j}} \sum\limits_{r \in T_i, s \in T_j}
Q_{rs} - A_{Q;rs}
\bigg)^2
\geq t^2
\bigg]\leq 2 \exp(-0.5 t^2 m)
\]
Setting $t = C \sqrt{\frac{\log n}{m}}$ for large enough $C = O(1)$, a union bound tells us that with probability $\geq 1 - n^{-8}$, the Bernoulli error is bounded by $t^2$. 
\end{proof}

\begin{cor}
The Bernoulli error is at most $O(\sqrt{\frac{\log n_Q}{n_Q}})$ with probability $\geq 1 - n_Q^{-4}$. 
\label{cor:bernoulli-error-row-dist}
\end{cor}

The rest of this section is devoted to bounded the smoothing errors $J_S(i, j)$.
% \subsubsection{Decomposing into Smoothing and Bernoulli Error}

% \]
% \subsubsection{Bernoulli Error}

% \blue{Todo - look at probability of landing in the $t$-thickening of the shell at $[0,1]^d$. It should actually be high by Talagrand, maybe.}

% If $\log n = O(\log n_Q)$ then this is exactly the $\sqrt{(\log n_Q) / n_Q}$ type bound that we expect. 

% \blue{todo calculation for $\eps_n$. 

\subsubsection{Latent distance to graph distance}\label{appendix:latent-to-graph-distance}

We claim that if nodes are close in the latent space then they are close in graph distance. 
% [Restatement of Proposition~\ref{prop:latent-to-graph-dist}]
\begin{prop}
Suppose that $\norm \bx_i - \bx_r \norm \leq \eps$ and $Q$ is $\beta$-smooth. Then $d_Q(i,r)\leq C_\beta^2 n^3 \eps^{2(\beta \land 1)}$. 	
\label{prop:latent-to-graph-dist}
\end{prop}
\begin{proof}
We the use smoothness of $Q$. By definition there exists $C_\beta>0$ such that $Q_{ki} - Q_{kr} \leq C_\beta \norm \bx_i - \bx_r \norm^{\beta \land 1}$. Therefore, 
\begin{align*}
d_Q(i,r) &:= \sum\limits_{\ell \neq i, r}\abs{(Q^2)_{\ell i} - (Q^2)_{\ell r}}^2 \\
&= \sum\limits_{\ell \neq i, r} \bigg(\sum\limits_{k \in [n]} Q_{\ell k}(Q_{k i} - Q_{kr})\bigg)^2 \\
&\leq \sum\limits_{\ell \neq i, r} \sum\limits_{k \in [n]} Q_{\ell k}^2 C_\beta^2 \eps^{2(\beta \land 1)} \\
&\leq n^3 C_\beta^2 \eps^{2 (\beta \land 1)}
\end{align*}	
\end{proof}

We can now bound the minimum sizes of the neighborhoods using the concentration of latent positions and smoothness of the graphon. 

\begin{lemma}[\cite{vershynin2018high}]
The volume of a ball of radius $r > 0$ in $\RR^d$ is $\frac{\sqrt{\pi}^d}{\Gamma(d/2 + 1)} r^d$, where $\Gamma(\cdot)$ is the $\Gamma$ function. 
\label{lemma:d-volume}
\end{lemma}

\begin{prop}
Let $C_d = (\Gamma(\frac{d}{2} + 1))^{1/d}$. Let $C_0, C^\prime$ be constants. If $\upsilon_n \geq C \cdot C_d (\sqrt{\frac{\log n}{n_Q}})^{1/d}$  for large enough constant $C > 0$, 
and $g_n = C_0 C_\beta^2 n^2 (\upsilon_n)^{2(\beta \land 1)}$, then with probability $\geq 1 - n^{-6}$ for all $i \in [n]$ the neighborhood size is $\abs{\{r: d_Q(i, r) \leq g_n\}} \geq C^\prime n_Q \sqrt{\frac{\log n}{n_Q}}$. 
\label{prop:lb-nbhd-size}
\end{prop}
\begin{proof}
Fix $i \in [n]$ and $\upsilon_n > 0$. Let $\eps_i$ denote the Lebesgue measure of $\mathrm{Ball}(\bx_i, \upsilon_n) \cap \X$. By Lemma~\ref{lemma:d-dim-ball} and Lemma~\ref{lemma:d-volume}, for all $i$, $\eps_i \geq (\frac{\sqrt{\pi} \upsilon_n}{2C_d})^d = (\frac{0.5 \sqrt{\pi} \upsilon_n}{C_d})^d$. Let $\eps = \min_{i \in [n]} \eps_i$.
%Let $\eps = (\frac{\sqrt{\pi} \upsilon_n}{(\Gamma(d/2 + 1))^{1/d}})^d$, so that $\eps \leq v_\eps^i$ for all $i$. 

By Corollary~\ref{cor:latent-conc-S-epsilon-ball}, with probability $\geq 1- n^{-8}$, there are $n_Q \eps - C \sqrt{\frac{\log n}{n_Q}}$ members $j$ of $S$ such that $\norm \bx_i - \bx_j \norm \leq \upsilon_n$. A union bound over $i$ gives the result simultaneously for all $i$ with probability $\geq 1- n^{-6}$. 

From Proposition~\ref{prop:latent-to-graph-dist}, it follows that for all $i \in [n]$, 
\[\abs{\{r \in S: d_Q(i,r) \leq C_\beta^2 n^2 (2 \upsilon_n^\prime)^{2(\beta \land 1)}\}} \geq n_Q \eps - 10 \sqrt{\frac{\log n}{n_Q}}
\] 
% Similarly, 
% \[\abs{\{r \in S: d_P(i,r) \leq C_\alpha^2 n^2 (2 \upsilon_n^\prime)^{2(\alpha \land 1)}\}} \geq n \eps - 10 \sqrt{\frac{\log n}{n_Q}}
% \]
Choosing $\upsilon_n \geq C \cdot C_d (\frac{\log n}{n_Q})^{\frac{1}{2d}}$ for large enough $C > 0$ gives the conclusion. 
\end{proof}
% \blue{notice taking the union bound over all $i \in [n]$ is what forces $\log n$ rather than $\log n_Q$. But if $n \leq n_Q^{O(1)}$ this is fine.}

\subsubsection{Graph Distance Concentration}
Next, we show that the empirical graph distance concentrates to the population distance.
\begin{prop}
For any arbitrary symmetric $P \in [0,1]^{n\times n}$, we have, for all $i, j$ simultaneously with probability at least $\geq 1 - O(n^{-8})$, that: 
\[
\abs{d_{A_P}(i, j) - d_P(i,j)} \leq O(n^2 \log n) + O
(n^{2.5} 
\sqrt{\log n})
\]	
\label{prop:graph-distance-conc-mcs}
\end{prop}
\begin{proof}
Fix $i, j$. Let $C_{ij} := (A_P^2)_{ij}$. By \cite{mcs21} A.1, we have $C_{ij} = (P^2)_{ij} + t_{ij}$ for an error term $t_{ij}$ such that $\PP[\forall i, j: \abs{t_{ij}} \leq 10 \sqrt{n \log n}] \geq 1 - n^{-10}$. Then, 
\begin{align*}
\abs{d_{A_P}(i, j) - d_P(i,j)} &= 
\abs{\sum\limits_{\ell \neq i, j} \big((C_{i\ell} - C_{j\ell})^2 - ((P^2)_{i \ell} - (P^2)_{j\ell})^2 \big)} \\
&= \sum\limits_{\ell \neq i, j} 
\abs{(t_{i\ell} + t_{j \ell})^2 + 2(t_{i\ell} + t_{j \ell}) ((P^2)_{i \ell} - (P^2)_{j\ell})} \\
&\leq O(n^2 \log n) + 
 O\bigg(
\sqrt{n \log n}\sum\limits_{\ell \neq i, j} \big((P^2)_{i \ell} - (P^2)_{j \ell} \big)
\bigg)
\end{align*}
Finally, notice that all entries of $P^2$ are of size $O(n)$, so the conclusion follows. 
\end{proof}

% \begin{prop}[Concentration of $A_P$-distance versus $P$-distance]
% There exists $c > 0$ such that with probability at least $1 - n^{-8}$, for all $i \in [n]$ and $r \in [n] \setminus \{i\}$ simultaneously, we have: 
% \[
% d_{A_P}(i, r) - 20 \sqrt{\frac{\log n}{n}}
% \leq d_P(i,r)
% \leq d_{A_P}(i, r) + 20 \sqrt{\frac{\log n}{n}}
% \]
% \label{prop:graph-distance-conc}
% \end{prop}
% \begin{proof}
% We first invoke the concentration bound Prop \ref{prop:entrywise-conc} with $\eps = C \sqrt{n \log n}$ for $C = 10$, then $\PP[\forall i \neq j: \abs{(A^2)_{ij} - (P^2)_{ij}} \leq \eps]\geq 1 - n_Q^{-6}$. 
% Then for $r \in T_i^{A_P}(h)$, it follows that
% \[
% \max\limits_{k \neq i, r} \abs{(e_i - e_r)^T P^2 e_k}
% \leq \max\limits_{k \neq i, r} \abs{(e_i - e_r)^T A_P^2 e_k} + 2 \eps = n d_{A_P}(i, r) + 2\eps
% \]
% Hence $d_P(i,r) \leq d_{A_P}(i, r) + 2C \sqrt{\frac{\log n}{n}}$. The other direction is similar. 
% \end{proof}

Finally, we will show that taking the restriction of the graph distance $T_i^P$ to nodes in $S \subset [n]$ does not incur too much error. 
% \blue{The $\Delta_n$ - insert the ref to the right assumpiton.}
\begin{prop}
Suppose $n = n_Q^{O(1)}$. Then there exists a constant $C$ such that if $h_0 \geq C \sqrt{\frac{\log n}{n_Q}} + \Delta_n$, then for all $i, r$ simultaneously, $r \in T_i^{A_P}(h_0)$ implies  $r \in T_i^{P}(h_2)$ for some $h_2 = O(h)$ with probability $\geq 1 - O(n^{-5})$. 
\label{prop:graph-distance-ap-p}
\end{prop}

\begin{proof}
Let us introduce the notation $T_i^{P, S}(h)$ to denote the bottom $h$-quantile of $\{d_P(i, j): j \in S\}$. In this notation, $T_i^{A_P}(h) := T_i^{A_P, S}(h)$ since we restrict the quantile to nodes in $S$. From Proposition~\ref{prop:graph-distance-conc-mcs} and Assumption~\ref{assumption:mcs}, we know that if $n \geq n_Q$ then for $h_0 \leq h_1 - 20 \sqrt{\frac{\log n}{n}} - \Delta_n$ we have $T_i^{A_P}(h_0) \subseteq T_i^{P, S}(h_1)$ simultaneously for all $i \in [n]$ with probability $\geq 1 - O(n^{-8})$. It remains to compare $T_i^{P, S}(h_1)$ with $T_i^{P}(h_2)$ for some $h_2$. 

We claim that if $h_2 \geq 30 \sqrt{\frac{\log n_Q}{n_Q}}$ then $\PP[\forall i \abs{T_i^{P} \cap S} \geq h_2 n_Q - 3 \sqrt{n_Q \log n_Q}] \geq 1 - O(n_Q^{-2})$. To see this, fix $i \in [n]$ and consider $T_i^P(h_2)$. For $j \in S$ let $X_j$ be the indicator variable: 
\[
X_j = \begin{cases}
1 & j \in T_i^{P}(h_2) \\
0 & \text{otherwise} 
\end{cases}
\]
Notice that $\abs{T_i^P(h_2) \cap S} = \sum_{j \in S} X_j$. By Hoeffding's inequality, since $\EE[\sum_{j \in S} X_j] = h_2 n_Q$ 
and $\abs{X_j - h_2} \leq 1$ for all $j$, we have: 
\begin{align*}
\PP[\abs{\abs{T_i^P(h_2) \cap S} - h_2 n_Q} \geq 3 \sqrt{n_Q \log n}]	\leq 2 \exp(-\frac{6 n_Q^2 \log n}{n_Q^2}) \leq 2n^{-6}
\end{align*}
Taking a union bound over all $i \in [n]$ shows the claim holds with probability $\geq 1 - O(n^{-5})$. Therefore we set $h_1 \leq h_2 - 3.1 \sqrt{\frac{\log n}{n_Q}}$ then $j \in T_i^{P, S}(h_1)$ implies $j \in T_i^P(h_2)$. 

The conclusion follows with $C = 24 \sqrt{\frac{\log n}{\log n_Q}} = O(1)$. 
\end{proof}

The ranking condition (Definition ~\ref{defn:ranks}) then allows us to translate between graph distances in $A_P$ and $Q$. 

\begin{cor}
Suppose that Definition~\ref{defn:ranks} holds for $(P, Q)$ at $h_n = c \sqrt{\frac{\log n_Q}{n_Q}} + \Delta_n$, for large enough constant $c > 0$. Suppose $n_Q \leq n \leq n_Q^{O(1)}$. Then for $h > h_n$ and $r \in T_i^{A_P}(h)$, it follows that $r \in T_i^{Q}(h_3)$ for some $h_3 = O(h)$. The statement holds simultaneously for all $i, r$ with probability $\geq 1 - O(n^{-5})$. 
\label{cor:graph-distance-ap-q}
\end{cor}

% \blue{note the bandwidth has dependence on $n_Q$ not $n$. Of course we need $n_Q \leq n$}

\subsubsection{Control of smoothing error}

We will decompose smoothing error into a sum of two terms called $E_{S, 1}$ and $E_{S, 2}$. The control of $E_{S, 1}$ is relatively straightforward. 

\begin{lemma}
The total smoothing error can be bounded with two terms: 
\[
\frac{2}{n^2} \sum\limits_{i, j \in [n]} J_S(i,j)
\leq 
E_{S, 1} + E_{S, 2}
\]
Where
\begin{align*}
E_{S, 1} := \frac{C}{n} \max\limits_{j \in [n], s \in T_j}\norm Q(e_j - e_s)\norm_2^2 \\
E_{S, 2} := \frac{4}{n^2} \sum\limits_{i \in [n]} \frac{1}{\abs{T_i}}
\EE \bigg[
\sum\limits_{r \in T_i} \sum\limits_{j \in [n]} \sum\limits_{s \in T_j}
(Q_{rj} - Q_{rs})^2
\bigg]
\end{align*}
\label{lemma:error-decomp}
\end{lemma}

\begin{proof}
\begin{align*}
\frac{2}{n^2} \sum\limits_{i, j \in [n]} J_S(i,j)
&= \frac{2}{n^2} \sum\limits_{i, j \in [n]} \frac{1}{\abs{T_i}^2 \abs{T_j}^2} \EE \bigg[
\bigg(\sum\limits_{r \in T_i, s \in T_j} Q_{ij} - Q_{rs}
\bigg)^2 
\bigg] \\
&\leq \frac{2}{n} \sum\limits_{i \in [n]} \frac{1}{n \abs{T_i}}
\sum\limits_{j \in [n]}
\frac{2}{\abs{T_j}}
\EE \bigg[\sum\limits_{r \in T_i, s \in T_j}
(Q_{ij} - Q_{rj})^2 + (Q_{rj} - Q_{rs})^2
\bigg] \\
&= \frac{4}{n} \sum\limits_{i \in [n]} \frac{1}{n \abs{T_i}}
\EE \bigg[
\sum\limits_j \frac{1}{\abs{T_j}}
\bigg(\sum\limits_{r \in T_i} (Q_{ij} - Q_{rj})^2 
+ \sum\limits_{r \in T_i} \sum\limits_{s \in T_j}
(Q_{rj} - Q_{rs})^2
\bigg)
\bigg]
\end{align*}
The second inner summand is precise $E_{S, 2}$. For $E_{S, 1}$, notice that $\abs{T_i} = \abs{T_j} = h(n_Q - 1)$ by definition. Therefore 
\[\sum_j \frac{1}{\abs{T_j}} \sum_{r \in T_i} (Q_{ij} - Q_{rj})^2 = \frac{1}{h(n_Q - 1)} \sum_{r \in T_i} \sum_j (Q_{ij} - Q_{rj})^2 \leq 2 \max_{r \in T_i} \norm (e_i - e_r)^T Q \norm_2^2
\]
\end{proof}

We can now bound $E_{S,1}$ in terms of graph distances.
\begin{lemma}
The smoothing error term $E_{S, 1}$ can be bounded as: 
\begin{align*}
E_{S, 1} &\leq \frac{2}{n} \max\limits_{i \in [n], r \in T_i} \sqrt{d_Q(i,r)} + \frac{2c}{\sqrt{n}}
\end{align*}
For constant $c > 0$.
\label{lemma:es1-to-graph-dist}
\end{lemma}
\begin{proof}
Fix $i \in [n]$ and $r \in T_i$. We have: 
\begin{align*}
\norm Q(e_i - e_r)\norm_2^2 
&\leq \norm e_i - e_r\norm_2 \norm Q^T Q (e_i - e_r)\norm_2 \\
&\leq 2\norm Q^2 (e_i - e_r)\norm_2 \\
\end{align*}

Now we will pass to graph distances. Let $e_{ab} := ((Q^2)_{aa} - (Q^2)_{ab})^2$ for $a, b\in [n]$. Notice that $\norm Q^2 (e_i - e_r)\norm_2 = \sqrt{d_Q(i, r) + e_{ir} + e_{ri}}$. Moreover, $\sqrt{e_{ir} + e_{ri}} \leq 2\sqrt{n}$ since the entries of $Q^2$ are individually bounded by $O(n)$. The conclusion follows. 
\end{proof}

% \label{todo fix cor ref.}
% \blue{todo shouldnt we have $h_n^2 n_Q^2$ factor?}
% By Proposition~\ref{prop:latent-to-graph-dist} we 
% obtain: 

\begin{prop}
Suppose $\Delta_n = O(\sqrt{\frac{\log n}{n_Q}})$. Let $C_d$ be the constant of Proposition~\ref{prop:lb-nbhd-size}. Then if the bandwidth of Algorithm~\ref{alg:q-averaging-row-wise} is $h_n = C \sqrt{\frac{\log n}{n_Q}}$, for a constant $C = O(1)$, then the smoothing error $E_{S, 1}$ is at most: 
\[
E_{S, 1} \leq 
C_2 C_d^{\beta \land 1} \sqrt{\frac{\log n_Q}{n_Q}}^{\frac{\beta \land 1}{d}}
\]
For some $C_2 = O(1)$, with probability $\geq 1 - O(n^{-6})$.
\label{prop:smoothing-error-1}
\end{prop}
\begin{proof}
Fix $i \in [n]$ and $r \in T_i^{A_P}(h_n)$. By Corollary~\ref{cor:graph-distance-ap-q}, if $h_n \geq C \sqrt{\frac{\log n}{n_Q}} + \Delta_n$ for a large enough constant $C > 0$, then there exists constant $C_2 > 0$ such that the following holds. With probability $\geq 1 - O(n^{-5})$, for all $i \in [n]$ and $r \in S$, $r \in T_i^Q(C_2 h_n)$, 

Let $\upsilon_n = C C_d (\sqrt{\frac{\log n}{n_Q}})^{1/d}$ for $C_d$ as in Proposition~\ref{prop:lb-nbhd-size} and $C > 0$ large enough constant. Then by Proposition~\ref{prop:lb-nbhd-size} the set of $s \in S$ such that $d_Q(i, r) \leq C_0 C_\beta^2 n^2 (\upsilon_n)^{2 (\beta \land 1)}$ has size at least $C_2 n_Q \sqrt{\frac{\log n}{n_Q}}$.The statement holds for all $i$ simultaneously with probability at least $1 - O(n^{-6})$. Therefore for all $i \in [n]$ and $r \in T_i^{A_P}(h_n)$, we have: 
\[
d_Q(i, r) \leq C_0 C_\beta^2 n^2 (\upsilon_n)^{2 (\beta \land 1)}
\]
For some $C_0, C_\beta = O(1)$, with probability $\geq 1 - O(n^{-6})$. By Lemma~\ref{lemma:es1-to-graph-dist} we  conclude that $E_{S,1}$ is bounded by $2 \upsilon_n^{\beta \land 1} + \frac{2}{\sqrt{n}}$ with the same probability. 
\end{proof}
% \blue{todo d dim volume stuff.}

% We simply need to show there are at least $(n_Q \eps_n^\prime) / 4$ nodes in $S$ that are distance $\eps_n^\prime$ apart from $i \in [n]$ in latent space, for each $i$. This follows Corollary~\ref{cor:latent-conc-S-epsilon-ball}. 

\subsubsection{Control of second smoothing error}

In this section we show that the second smoothing error can be controlled in terms of $E_{S, 1}$. We will need to track the following quantity.

\begin{defn}[Membership count]
For $r \in S$ and bandwidth $h$, distance cutoff $\eps$, the $P$-neighborhood count of $r$ is $\psi_P(r) := \abs{\{j \in [n]: r \in T_j^P(h, \eps)\}}$.
\end{defn}
In words, $\psi_P(r)$ counts the number of nodes $j \in [n]$ such that $r$ lands in the neighborhood of $j$ in our algorithm. While we know that $\abs{T_j^P(h)} \leq hn_Q$ always, simply applying the pigeonhole principle gives too weak of a bound on membership counts. The base case is that there may be a ``hub'' node $r$ lands in $T_j^P(h)$ for all $j$. We will show that there can be no such hub node. 

Supposing that we can control of the empirical count $\psi_{A_P}$, we show that the smoothing error can be bounded.
\begin{prop}
Let $h_n$ be the bandwidth. Then: 
\[
E_{S, 2} \leq O(\frac{E_{S, 1}}{h_n n}) \cdot \max\limits_{r \in [n]} (\psi_{A_P}(r))
\]
\label{prop:es2-via-es1}
\end{prop}

\begin{proof}
Rearranging terms, we have: 
\begin{align*}
E_{S, 2} &= \frac{1}{n^2 h^2 n_Q^2} \sum\limits_{i, j \in [n], r \in T_i, s \in T_j} (Q_{rj} - Q_{rs})^2 \\	
&= \frac{1}{n^2 h^2 n_Q^2} \sum\limits_{r \in S} \psi_{A_P}(r) \sum\limits_{j, s} (Q_{rj} - Q_{rs})^2 \\
&= \frac{n_Q}{n^2 h^2 n_Q^2} \EE\limits_{r \in S}\bigg[
\psi_{A_P}(r) \sum\limits_{j, s} (Q_{rj} - Q_{rs})^2
\bigg] \\
&= \frac{n_Q}{n^2 h^2 n_Q^2} \EE\limits_{r \in [n]}\bigg[
\psi_{A_P}(r) \sum\limits_{j, s} (Q_{rj} - Q_{rs})^2
\bigg] 
\end{align*}
Where the last step is because $j, s$ do not depend on $i, r$ and because $S \subset [n]$ is chosen uniformly at random. Now, we will control the expectation by passing to a row sum, which is handled by $E_{S, 1}$. 
\begin{align*}
 \EE\limits_{r \in [n]}\bigg[
\psi_{A_P}(r) \sum\limits_{j, s} (Q_{rj} - Q_{rs})^2
\bigg] 
&\leq \max\limits_{r \in [n]} \bigg(\frac{\psi_{A_P}(r)}{n}\bigg) \cdot \sum\limits_{j \in [n]}\sum\limits_{s \in T_j}
\norm Q(e_j - e_s)\norm_2^2	
\end{align*}
Recall that $n^2 n_Q h_n E_{S, 1} = \Omega(\sum\limits_{j \in [n]}\sum\limits_{s \in T_j}
\norm Q(e_j - e_s)\norm_2^2)$. Hence we conclude that: 
\[
E_{S, 2} \leq O(\frac{E_{S, 1}}{h_n n}) \cdot \max\limits_{r \in [n]} (\psi_{A_P}(r))
\]
\end{proof}

We therefore must show that $\max\limits_{r \in S} \psi_{A_P}(r) \leq O(h n)$ with high probability.
% the Assumption 3.2 of \cite{mcs21}  

% \blue{Track the $2^d$ or whatever constants here.}
\begin{prop}[Population Version]
Suppose Assumption~\ref{assumption:mcs} holds for $P$ with $c_1 < c_2$ and $\Delta_n = O((\frac{\log n}{n_Q})^{\frac{1}{2} \lor \frac{\alpha \land 1}{d}})$. Then if $h \leq C \sqrt{\frac{\log n}{n_Q}}$ for large enough constant $C >0$, then we have $\max\limits_{r \in S} \psi_{P}(r) \leq O(h n)$ with probability at least $1 - O(n_Q^{-8})$.
\label{prop:psi-bound-population}
\end{prop}
\begin{proof}
Fix $r \in S$. Let $C_d$ be as in Proposition~\ref{prop:lb-nbhd-size}.Suppose that $\eps = C_d (C + 10) \sqrt{\frac{\log n_Q}{n_Q}}^{1/d}$	and 
$h = C \sqrt{\frac{\log n_Q}{n_Q}}$. Now, we will claim that for large enough constant 
$c > 0$, that $\psi_P(r)$ is at most the size of $\mathrm{Ball}(\bx_r, c \eps) \cap \{\bx_1, \dots, \bx_{n}\}$. 

Suppose that $c > 0$ is a large enough constant. Now suppose that $\bx_j$ is such that $\norm \bx_j - \bx_r \norm \geq c \eps$. We can lower bound the graph distance using Assumption~\ref{assumption:mcs}, as: 
\[
d_P(r, j) := \norm (e_r - e_j)^T P^2 (I - e_r e_r^T - e_j e_j^T)\norm_2^2 
\geq c_1 n^3 (c \eps)^{2(\alpha \land 1)} 
- n^3 \Delta_n
\]
On the other hand, suppose that $i \in S$ is such that $\norm \bx_i - \bx_j \norm \leq \eps$. Then $d_P(i, j) \leq C_\alpha^2 n^3 \eps^{2(\alpha \land 1)}$ by Proposition~\ref{prop:latent-to-graph-dist}. Therefore since $\eps = C_d (C + 10) \sqrt{\frac{\log n_Q}{n_Q}}^{1/d}$ and $\Delta_n = O((\frac{\log n}{n_Q})^{\frac{1}{2} \lor \frac{\alpha \land 1}{d}})$, for large enough $c_1 > 0$ we have: 
\[
d_P(r, j) := \norm (e_r - e_j)^T P^2 (I - e_r e_r^T - e_j e_j^T)\norm_2^2 
\geq \frac{c_1}{2} n^3 (c \eps)^{2(\alpha \land 1)} 
\]
Then, if we choose $c > 0$ such that $c^{2(\alpha \land 1)} > \frac{2 C_\alpha^2}{c_1}$, then $d_P(i, j) < d_P(r, j)$. 

Next, from our choices of $h, \eps$, by Corollary~\ref{cor:latent-conc-S-epsilon-ball}, simultaneously for all $i \in [n]$ there are at least $h n_Q$ nodes in $S$ that have distance $\leq \eps$ in latent space from $\bx_i$, with probablity $\geq 1 - O(n_Q^{-6})$. 

Therefore, if $\bx_r \not \in \mathrm{Ball}(\bx_j, c \eps) \cap \{\bx_1, \dots, \bx_{n}\}$ 
then $r \not \in T_j^P(h)$. This implies that $\psi_P(r) \leq \abs{ \{\mathrm{Ball}(\bx_r, 2 c \eps) \cap \{\bx_1, \dots, \bx_{n}\}}$. We can bound the size of this ball with Lemma~\ref{lemma:latent-conc-d-dim}. Notice the Lebesgue measure of $\mathrm{Ball}(\bx_r, 2 c \eps) \cap [0,1]$ is at most $(\frac{4c\eps}{C_d})^d$. Therefore, since $\bx_i$ are chosen iid from the Lebesgue measure on $\mathcal{X}$, with probability at least $\geq 1 - O(n_Q^{-10})$, we have: 
\[
\frac{1}{n} \abs{\mathrm{Ball}(\bx_r, 2 c \eps) \cap \{\bx_1, \dots, \bx_{n}\}} \leq 2c\eps + 10 \sqrt{\frac{\log n}{n}}
\]
The right-hand side is bounded by $O(h)$ if $n \geq n_Q$. Taking a union bound over all $r \in S$ gives the conclusion. 
\end{proof}
% \blue{todo track and correct the $d$ factors.}

We conclude with the desired upper bound. 
\begin{prop}[Bound on $\psi_{A_P}(r)$]
Suppose Assumption~\ref{assumption:mcs} holds for $P$ with $c_1 < c_2$ and $\Delta_n = O((\frac{\log n}{n_Q})^{\frac{1}{2} \lor \frac{\alpha \land 1}{d}})$. Then if $h \leq C_0 \sqrt{\frac{\log n_Q}{n_Q}}$ for small enough constant $C_0$, then we have $\max\limits_{r \in S} \psi_{A_P}(r) \leq O(h n)$ with probability at least $1 - O(n_Q^{-8})$.
\label{prop:psi-ap-bound}
\end{prop}

\begin{proof}
By Proposition~\ref{prop:graph-distance-conc-mcs}, with probability at least $1 - O(n_Q^{-8})$, we have for all $r \in S, j \in [n]$ simultaneously that: 
\begin{align*}
d_{A_P}(r, j) &\geq d_P(r, j) - O(n^{2.5} \sqrt{\log n}) \\
&\geq (1 - O(\frac{1}{\sqrt{n}})) d_P(r, j)
\end{align*}

Similarly $d_{A_P}(r, j) \leq (1 + O(\frac{1}{\sqrt{n}})) d_P(r, j)$. We conclude that $\psi_{A_P}(r) \leq 2 \psi_{P}(r) = O(h n)$ with probability $\geq 1 - O(n_Q^{-8})$. 
\end{proof}

\subsubsection{Overall Error}

We can bound $C_d := \Gamma(\frac{d}{2} + 1)^{1/d}$ with the elementary inequality. 
\begin{lemma}
Let $C_d := \Gamma(\frac{d}{2} + 1)^{1/d}$. Then $C_d \leq \sqrt{d/2}$. 
\label{lemma:dim-const-bound}
\end{lemma}

\begin{proof}[Proof of Theorem~\ref{thm:alg1error}]
By Proposition~\ref{prop:psi-ap-bound} and Prop~\ref{prop:es2-via-es1}, we have that $E_{S, 1} \leq O(E_{S, 1})$ with probability at least $1 - O(n_Q^{-8})$. Therefore by Proposition~\ref{prop:smoothing-error-1}, 
\[
\PP \bigg[E_{S, 1} + E_{S, 2} 
\leq O\bigg(
C_d^{\beta \land 1}
\bigg(\frac{\log n}{n_Q}\bigg)^{\frac{\beta \land 1}{2d}}\bigg)\bigg] \geq 1 - O(n_Q^{-6})
\]
By Lemma~\ref{lemma:dim-const-bound}, $C_d \leq \sqrt{d/2}$. Finally, by Corollary~\ref{cor:bernoulli-error-row-dist}, the Bernoulli error is bounded by $O(\sqrt{\frac{\log n_Q}{n_Q}})$ with probability $\geq 1 - O(n_Q^{-4})$. Applying a union bound over the two kinds of error and Lemma~\ref{lemma:error-decomp} gives the result.
\end{proof}

% !TEX root = ./neurips_2024.tex
% \section{Appendix / supplemental material}

% Optionally include supplemental material (complete proofs, additional experiments and plots) in appendix.
% All such materials \textbf{SHOULD be included in the main submission.}

\subsection{Proof of Theorem~\ref{thrm:minimax-sbm}}

Recall the Gilbert-Varshamov code \citep{guruswami-book-coding-theory}. 
\begin{theorem}[Gilbert-Varshamov]
Let $q \geq 2$ be a prime power. For $0 < \eps < \frac{q-1}{q}$ there exists an $\eps$-balanced code $C \subset \FF_q^n$ with rate $\Omega(\eps^2 n)$. 
\label{thrm:gv-code}
\end{theorem}
% \begin{proof}
% With high probability a random linear code satisfies these conditions. 
% \end{proof}

We will use the following version of Fano's inequality.
\begin{theorem}[Generalized Fano Method, \cite{yu-1997}]
Let $\mathcal{P}$ be a family of probability measures, $(\mathcal{D}, d)$ a pseudo-metric space, and $\theta: \mathcal{P} \to \mathcal{D}$ a map that extracts the parameters of interest. For a distinguished $P \in \mathcal{P}$, let $X \sim P$ be the data and $\hat \theta := \hat \theta(X)$ be an estimator for $\theta(P)$. 

Let $r \geq 2$ and $\mathcal{P}_r \subset \mathcal{P}$ be a finite hypothesis class of size $r$. Let $\alpha_r, \beta_r > 0$ be such that for all $i \neq j$, and all $P_i, P_j \in \mathcal{P}_r$, 
\begin{align*}
d(\theta(P_i), \theta(P_j)) &\geq \alpha_r \\
KL(P_i, P_j) &\leq \beta_r
\end{align*}
Then, 
\begin{align*}
\max\limits_{j \in [r]} \EE\limits_{P_j} d(\hat \theta(X), \theta(P_j)) &\geq \frac{\alpha_r}{2}
\bigg(
1 - \frac{\beta_r + \log 2}{\log r}
\bigg)	
\end{align*}
\label{thrm:yu-fano}
\end{theorem}

\begin{defn}[Relative Hamming Distance]
For $\bx, \by \in \{0,1\}^m$ the relative Hamming distance is: 
\[
d_H(\bx, \by) := \frac 1 m \abs{\{i \in [m]: x_i \neq y_i\}}
\]	
\end{defn}

We will need the following construction of coupled codes. 
\begin{prop}
Let $m_P, m_Q \geq 2$ and $m_Q$ divide $m_P$. There exists a code $C \subset \{0,1\}^{m_P}$ and a projection map $\Pi: \{0,1\}^{m_P} \to \{0,1\}^{m_Q}$ such that if $C^\prime = \{\Pi(w): w \in C\}$ then $C^\prime$ is a code with relative Hamming distance $\Omega(1)$. Moreover, $\abs{C} = \abs{C^\prime} \geq 2^{0.1 m_Q}$
\label{prop:code-coupled}
\end{prop}

Throughout the proof, we will identify the community assignment function $z: [n] \to [k]$ of an SBM (Definition~\ref{defn:sbm-main}) with the matrix $Z \in \{0,1\}^{n \times k}$ where $Z_{ij} = 1$ if and only if $z(i) = j$. 

\begin{proof}
Begin with a Gilbert-Varshamov code $B \subset \{0,1\}^{m_Q}$  as in Theorem~\ref{thrm:gv-code}. We can ``lift'' $B$ to a code on $\{0,1\}^{m_P}$ simply by concatenation.
% \marginpar{\blue{AM: to a code on $\{0,1\}^{m_P}$}}. 
If $w \in B$, then the corresponding $w^\prime \in C$ is just $w^\prime = (w, w, \dots, w) \in \{0,1\}^{m_P}$. Let $\Pi: \{0,1\}^{m_P} \to \{0,1\}^{m_Q}$ simply select the first $m_Q$ bits of a word. It is clear that $B = \{\Pi(w): w \in C\}$, so we are done.
\end{proof}
%  \blue{AM: this is true only when $m_P/m_Q \le$ constant. }
% \blue{We don't need to analyze the distance or rate properties of $C$, actually, only $C^\prime$. Cleaner to do that, I think.}

Now we are ready to prove Theorem~\ref{thrm:minimax-sbm}.

\begin{proof}[Proof of Theorem~\ref{thrm:minimax-sbm}]
Let $m_P = \binom{n}{2}$, $m_Q = \binom{n_Q}{2}$, and $m = m_P$. Let $C \subset \{0,1\}^{m_P}$ be the code and $\Pi: \{0,1\}^{m_P} \to \{0,1\}^{m_Q}$ the projection map of Prop~\ref{prop:code-coupled}. For each $w \in C$, we construct a pair of SBMs $P_w, Q_w \in \RR^{n \times n}$ as follows. 

Each $P_w, Q_w$ is a stochastic block model with $k_P, k_Q$ classes respectively. All the $P_w$ share the same community structure, namely the lexicographic assignment where nodes $1, 2, \dots, \frac{n}{k_P}$ are assigned to community $1$, and so on. Similarly all the $Q_w$ share the same lexicographic community structure with nodes $1, 2, \dots, \frac{n}{k_Q}$ assigned to community $1$, and so on. Therefore, there are fixed $Z_P \in \{0, 1\}^{n \times k_P}, Z_Q \in \{0, 1\}^{n \times k_Q}$, such that for all $w \in C$, there exist $A_w \in \RR^{k_P \times k_P}, B_w \in \RR^{k_Q \times k_Q}$ with: 
\begin{align*}
P_w = Z_P A_w Z_P^T \\
Q_w = Z_Q B_w Z_Q^T 
\end{align*}

The $A_w, B_w$ are defined as follows. Let $i, j \in [k_P]$ and $i^\prime, j^\prime \in [k_Q]$ be such that $i < j$ and $i^\prime < j^\prime$. Since $m_P = \binom{k_P}{2}$ and $m_Q = \binom{k_Q}{2}$, we can identify $(i,j)$ and $(i^\prime, j^\prime)$ with indices of $[m_P], [m_Q]$ respectively. Then for fixed $\delta_P, \delta_Q > 0$, the edge connectivity probabilities are: 
\begin{align*}
A_w(i, j) = A_w(j,i) := \begin{cases}
1/2 & w_{ij} = 0 \\
1/2 + \delta_P & w_{ij} = 1
\end{cases} \\
B_w(i^\prime, j^\prime) 
= B_w(j^\prime, i^\prime)
:= \begin{cases}
1/2 & \Pi(w)_{i^\prime j^\prime} = 0 \\
1/2 + \delta_Q & \Pi(w)_{i^\prime j^\prime} = 1
\end{cases} 
\end{align*}
We can set the diagonals of $A_w, B_w$ to be $1/2$ as well. 

Next, let $\mathcal{P}_r$ be a family of $r = \abs{C}$ probability measures. For fixed $w \in C$, the corresponding measure is the distribution over data $(A_P, A_Q) \in \{0,1\}^{n\times n} \times \{0,1\}^{n_Q \times n_Q}$ sampled from $(P_w, Q_w[S, S])$. Note that we restrict $S$ to be a fixed subset of $[n]$. 

Next, let $\theta((P_w, Q_w)) := Q_w$, and let $d(\theta((P_w, Q_w)), \theta((P_{w^\prime}, Q_{w^\prime}))) := \frac{1}{n} \norm Q_w - Q_{w^\prime} \norm_F$. We will show that for all $w, w^\prime \in C$ such that $w \neq w^\prime$ that: 
\begin{align*}
KL((P_w, Q_w), (P_{w^\prime}, Q_{w^\prime}))
&\leq KL(P_w, P_{w^\prime}) + KL(Q_w, Q_{w^\prime}) \\
&\leq O(n^2 \delta_P^2 + n_Q^2 \delta_Q^2) \\
&=: \beta \\
d((P_w, Q_w), (P_{w^\prime}, Q_{w^\prime}))
&:= \frac{1}{n} \norm Q_w - Q_{w^\prime} \norm_F \\
&\geq \Omega(\delta_Q) \\
&=: \alpha 
\end{align*}

For the $\beta$ claim, by Proposition 4.2 of \cite{gao2015rate}, if $\delta_P, \delta_Q \in (0, 1/4)$, we have: 
\begin{align*}
KL((P_w, Q_w), (P_{w^\prime}, Q_{w^\prime}))
&\leq KL(P_w, P_{w^\prime}) + KL(Q_w, Q_{w^\prime}) \\
&\lesssim \sum\limits_{i, j \in [n]} (P_w(i,j) - P_{w^\prime}(i,j))^2 + (Q_w(i,j) - Q_{w^\prime}(i,j))^2 
\end{align*}
% \blue{`AM: there se'ems to be a typo in first line. Also shouldn't the probabilities be within [1/2,3/4) for this?}

Next, notice that $A_w(i,j) \neq A_{w^\prime}(i,j)$ iff $w_{ij} \neq w_{ij}^\prime$. Then for distinct $w, w^\prime \in C$, we have $d_H(w, w^\prime) = \Omega(m_P)$, so 
\[\sum\limits_{i, j \in [n]} (P_w(i,j) - P_{w^\prime}(i,j))^2 
\lesssim \delta_P^2 \frac{n^2}{k_P^2}d_H(w, w^\prime) \binom{k_P}{2}
\lesssim \delta_P^2 n^2
\]
The bound for $Q_w$ is similar, so this verifies the $\beta$ claim. 

Similarly, for the $\alpha$ claim, notice that 

\[\frac{1}{n} \norm Q_w - Q_{w^\prime} \norm_F \gtrsim \frac{1}{k_Q} \sqrt{\delta_Q^2 d_H(\Pi(w), \Pi(w^\prime))} 
\geq \frac{\delta_Q}{k_Q} \sqrt{d_H(\Pi(w), \Pi(w^\prime))} 
\]
By Prop~\ref{prop:code-coupled}, $d_H(\Pi(w), \Pi(w^\prime)) = \Omega(m_Q) = \Omega(k_Q^2)$. Therefore $\alpha \leq \Omega(\delta_Q)$. 

Next, by Prop~\ref{prop:minimax-satisfies-assumption}, the pair $(P_w, Q_w)$ satisfies Definition~\ref{defn:ranks} for all $w \in C$. Moreover, $\log \abs{C} \geq 0.1 m_Q$ by Prop~\ref{prop:code-coupled}. 

Combining these results, by Theorem~\ref{thrm:yu-fano} the overall lower bound is: 
\begin{align*}
\inf\limits_{\hat Q} \sup\limits_{w} \frac{1}{n} \norm \hat Q - Q_w \norm_F 
&\gtrsim
\alpha \bigg(1 - \frac{\beta + \log 2}{0.1 \binom{k_Q}{2}} \bigg) \\
&\geq \delta_Q \bigg(1 - \frac{30 n^2 \delta_P^2}{k_Q^2} - \frac{30 n_Q^2 \delta_Q^2}{k_Q^2} - o(1)\bigg)
\end{align*}

If we choose $\delta_P = 0.01 (\frac{k_Q}{n})$ and $\delta_Q = 0.01\frac{k_Q}{n_Q}$, then:
\begin{align*}
\inf\limits_{\hat Q} \sup\limits_{w} \frac{1}{n^2} \norm \hat Q - Q_w \norm_F^2 
&\gtrsim \delta_Q^2 \\
&\gtrsim \frac{k_Q^2}{n_Q^2} 
\end{align*}
Note that $k_Q \leq n_Q \leq n$, so $\delta_P, \delta_Q \in (0,1/4)$ as desired. 
\end{proof}

% \blue{todo rewrite in terms of the ``at''}
\begin{prop}
If $h_{n} = \min\{\frac{1}{k_P}, \frac{1}{k_Q}\}$ then for all $w \in C$, the pair $(P_w, Q_w)$ satisfies Defn~\ref{defn:ranks} at $h_{n}$. 
\label{prop:minimax-satisfies-assumption}
\end{prop}

\begin{proof}
Consider $h = h_{n}$ and some node $i \in [n]$. Suppose that $j \neq i$ is in the same $P_w$-community as $i$, and that $\ell \neq i$ is in a different community. Then notice that $d_{P_w}(i, \ell) \geq d_{P_w}(i, j)$. Therefore $j \in T_{i}^{P_w}(h)$. Moreover, since $h \leq \frac{1}{k_P}$ and since the nodes of $S \subset [n]$ are equidistributed among the communities $1, 2, \dots, k_{P}$, it follows that all members of $T_{i}^{P_w}(h)$ must belong to the same $P_w$-community as $i$. 

Therefore, since the communities of $Q_w$ are a coarsening of the communities of $P_w$, $j \in T_i^{Q_w}(\frac{1}{k_Q})$. Since $h \leq \frac{1}{k_Q}$, we are done. 
\end{proof}

\subsection{SBM Clustering Error}\label{subsection:sbm-clustering-error}

In this section, we prove a minimax lower bound in the clustering regime for stochastic block models. 

% \blue{TODO fix this later.}

% \blue{We can show a lower bound even when $Q$ coarsens $P$.}

\begin{theorem}
Let $\Pi$ denote the parameter space of pairs of SBMs $(P, Q)$ on $n$ nodes with $k_P, k_Q$ communities respectively, such that the cluster structure of $Q$ is a coarsening the cluster structure of $P$. Then: 
\[
\inf\limits_{\hat Q} \sup\limits_{(P, Q) \in \Pi} \EE[\frac{1}{n^2} \norm \hat Q - Q_i \norm_F^2] 
\gtrsim \frac{\log k_Q}{n_Q}
\]	
\label{thrm:sbm-clustering-error}
\end{theorem}

\begin{proof}
Let $H_m \in [0,1]^{m \times m}$ be the Hadamard matrix of order $m$ modified to replace all entries $-1$ with $0$. If $m$ is not a power of two, let $H_m$ be defined as follows. Let $\ell = \floor*{\log_{2} m}$ and let $H_{m^\prime} \in \RR^{m/2 \times m/2}$ contain $H_{2^{\ell - 1}}$ on its top left block and zeroes elsewhere. Let: 
\[
H_m = \begin{bmatrix} 
\bm{0} \bm{0}^T & H_{m^\prime} \\
H_{m^\prime}^T & \bm{0} \bm{0}^T
\end{bmatrix}
\]
Notice that at most $\frac{7}{8}$ fraction of the entries of $H_m$ are zero-padded, for any $m$. Now, let $B_P = \frac 1 2 \bm{1}\bm{1}^T + \delta_P H_{k_P}$ and $B_Q = \frac 1 2 \bm{1}\bm{1}^T + \delta_Q H_{k_Q}$ for some $\delta_P, \delta_Q \in (0, 1/4)$ to be chosen later.

We will define two families of matrices indexed by a finite set $T$. For $i \in T$, there are some $Z_i \in \{0,1\}^{n \times k_P}$ and $Y_i \in  \{0,1\}^{n \times k_Q}$ to be specified later. Then, 
\begin{align*}
P_i &= Z_i B_P Z_i^T \\
Q_i &= Y_i B_Q Y_i^T
\end{align*}
Now, we define $Y_i$ as follows. Let $Z_{n, k_Q}$ denote the set of balanced clusterings $z: [n] \to [k_Q]$ such that for all $i, j \in [k_Q]$, $\abs{z^{-1}(\{i\})} = \abs{z^{-1}(\{j\})}$. Let $Z \subset Z_{n, k_Q}$ select the $z$ such that for all $j \leq k_Q / 2$, $z^{-1}(j) = \{\floor*{\frac{n(j-1)}{k_Q}}, \dots, \floor*{\frac{n j}{k_Q}}\}$. Define a distance on $Z$ as follows. For $y, y^\prime \in Z$ let $Y, Y^\prime \in \{0,1\}^{n \times k_Q}$ be the corresponding cluster matrices and let $d(y, y^\prime) := \frac{1}{n} \norm Y B_Q Y^T - Y^\prime B_Q (Y^\prime)^T \norm_F$. By Theorem 2.2 of \cite{gao2015rate}, there exists a packing $T_0 \subset Z$ with respect to $d$ such that for all $y, y^\prime \in T_0$, we have $\abs{\{j: y^\prime(j) \neq y(j)\}} \geq n / 6$. Moreover, $\log \abs{T_0} \geq \frac{1}{12} n \log k_Q$.
Set $T = T_0$. For any $y_i \in T_0$, let $Y_i \in \{0,1\}^{n \times k_Q}$ be the corresponding cluster matrix and then $Q_i = Y_i B_Q Y_i^T$. 

Now, to define $Z_i$, take $a \in [k_Q]$ and partition $y_i^{-1}(\{a\}) \subset [n]$ into $k_P / k_Q$ equally sized communities in a uniformly random way. Number these $1, \dots, \frac{k_P}{k_Q}$. In this way, we split community $1$ of $y_i$ into communities $1, \dots, \frac{k_P}{k_Q}$ of $z_i$, and so on. Define $Z_i$ to be the matrix corresponding to $z_i$. Notice that $Z_i, Y_i$ are both balanced clusterings and that the clustering $Y_i$ coarsens that of $Z_i$. Therefore $(P_i, Q_i)$ are a pair of heterogeneous symmetric SBMs satisfying Definition~\ref{defn:ranks} at $h = 1/k_Q$. 

Next, we apply Fano's Inequality (Theorem~\ref{thrm:yu-fano}). Recall $\log \abs{T} \geq \frac{1}{12} n \log k_Q$. Now, for $i, j \in T$ distinct, Prop 4.2 of \cite{gao2015rate} gives: 
\[
D_{KL}((P_i, Q_i), (P_j, Q_j))
\leq D_{KL}(P_i, P_j) + D_{KL}(Q_i, Q_j) 
\leq O(n^2 \delta_P^2 + n_Q^2 \delta_Q^2) =: \gamma_1
\]
Finally, we can bound: 
\begin{align*}
\frac{1}{n^2} \norm Q_i - Q_{i^\prime}\norm_F^2 &\geq 
\frac{1}{n^2} \sum\limits_{n / 2 < j \leq n} \frac{n}{k_Q} \norm (e_{y_i(j)} - e_{y_i^\prime(j)}) B_Q \norm^2 \\
&\geq c_0 \delta_Q^2 =: \gamma_2^2
\end{align*}
Where $c_0 > 1$ is some constant. This follows because there are a constant fraction of $j > n/2$ such that $y_i(j) \neq y_i^\prime(j)$, and any two rows of the Hadamard matrix differ on half their entries. 
% \begin{align*}
% \frac{1}{n^2} \norm Q_i - Q_{i^\prime} \norm_F^2 
% \geq \frac{1}{n^2} \sum\limits_{n/2 < j \leq n__P} x
% \end{align*}

Now, set $\delta_Q^2 = \frac{n_Q \log k_Q}{10 n_Q^2}$ and $\delta_P^2 = \frac{\log k_Q}{10 n^2}$. Since $n \geq n_Q$, we conclude that: 
\begin{align*}
\inf\limits_{\hat Q} \sup\limits_{i \in T} \EE[\frac{1}{n^2} \norm \hat Q - Q_i \norm_F^2] 
&\gtrsim \gamma_2^2 \bigg(1 - \frac{\gamma_1 + \log 2}{(1/12) n \log k_Q}\bigg) \\
&\gtrsim \frac{\log k_Q}{n_Q}
\end{align*}
\end{proof}

% \blue{Hadamard requires power of two, should rewrite to be more general.}

% \blue{AJ: We need to check that this condition on the $\eps$-cover of the set of clusters does not incur a dependence somewhere else. Gao says, above the equation for $\abs{B(z, \eps)}$ that choosing $\eps^2 = \frac{c_2 \log k}{n}$ works. It's a calculation...}

% \blue{We remark that we used nothing about the cluster structure of $P$ here, or really anything at all about $P$. All we need is for $P$ to be coupled. It can even have $k_P = k_Q$ communities.}

% \blue{Can we set $\delta_Q^2 = \frac{1}{10} \land \frac{n \log k_Q}{10 n_Q^2}$ here? Seems weird to do so, but I don't see where the proof would disallow it.}

% % {\bf Condition number of connectivity matrices.} Notice that condition number of the connectivity matrices of our assumption is equal to $1$. 

% \begin{prop}[\cite{yarlagadda1982note}]
% Let $H_n \in \RR^{2^n \times 2^n}$ be a Hadamard matrix of order $n$. Then $H_n$ has eigenvalues $\{2^{n/2}, - 2^{n/2}\}$ with multiplicity $2^{n-1}$ each. 
% \end{prop}

% \begin{cor}
% All singular values of $H_n$ equal $2^{n/2}$.
% \end{cor}
% This can also be seen from $H_n^T H_n = 2^n I_{2^n}$. 

% As a corollary, using $\sigma_{min}(AB) \geq \sigma_{min}(A) \sigma_{min}(B)$, we can lower bound the min singular values of $P_i, Q_i$ and also upper bound them. Then we should obtain Assumption~\ref{assumption:mcs}.
% \blue{Note that we replace the $-1$ with $0$ in the Hadamard matrices, although I think this step is unnecessary. We can actually use them straight up.}

\subsection{Proof of Proposition~\ref{prop:q-perfect-clustering}}

We first argue that Algorithm~\ref{alg:q-perfect-clustering} perfectly recovers $Z_P, Z_Q$ with high probability. 

\begin{theorem}[Implicit in \cite{chen2014improved}]
Let $M = ZBZ^T$ be an $(n, n_{min}, s)$-HSBM. Then there exists absolute constant $C > 0$ such that the Algorithm of \cite{chen2014improved} can recover $Z$, up to permutation, with zero error with probability $\geq 1 - O(n^{-8})$ if:
\[
s \geq C (\frac{\sqrt{n}}{n_{min}} \lor \frac{\log^2(n)}{\sqrt{n_{min}}})
\]
\label{thrm:chen-cluster}
\end{theorem}

\begin{proof}
The algorithm of \cite{chen2014improved} returns a matrix $Y \in \{0,1\}^{n \times n}$ such that $Y_{ij} = 1$ iff $i, j$ are in the same community, with probability $\geq 1 -O(n^{-8})$. Therefore, to construct a clutering from $Y$, simply assign the cluster of node $1$ to all $j \in [n]$ such that $Y_{1j} = 1$, and so on. This returns the true $Z \in \{0,1\}^{n \times k}$ up to permutation with probability $\geq 1 -O(n^{-8})$. Note that $k$ is correctly chosen because $Y$ is equal to a block-diagonal matrix of ones up to permutation, with $k$ blocks. 
\end{proof}

Theorem~\ref{thrm:chen-cluster} implies the following. 
\begin{prop}
Let $\hat Z_P, \hat Z_Q$ be as in Algorithm~\ref{alg:q-perfect-clustering}. Let $s_P, s_Q$ be the signal to noise ratios of $P, Q$ respectively. If $s_P, s_Q$ satisfy the conditions of Theorem~\ref{thrm:chen-cluster} with respect to $(n, n_{min}^{(P)}$ and $(n_Q,  n_{min}^{(Q)})$ respectively, then then with probability $\geq 1 - O(n_Q^{-8})$, there are permutation matrices $U_P \in \{0,1\}^{k_P \times k_P}, U_Q \in \{0,1\}^{k_Q \times k_Q}$ such that $\hat Z_P = Z_P U_P$ and $\hat Z_Q = Z_Q U_Q$. 
\label{prop:perfect-clustering-assortative}
\end{prop}

Next, we want to recover the clustering of $Q$ on all $n$ nodes, not just the $n_Q$ nodes that we observe in $A_Q$. This is given by the following.

% Let $I_S \in \{0,1\}^{n \times n}$ equal $1$ on the diagonal entries corresponding to $S$ and zero elsewhere. Then if $\Pi \in \{0,1\}^{k_P \times k_Q}$ is the true projection, we can recover $\Pi$ as follows. 

% Then if $\Pi \in \{0,1\}^{k_P \times k_Q}$ is the true projection, we can recover $\Pi$ as follows. 

% \blue{Need to handle the issue of $Q$ being a coarsening of $P$ vis a vis Assumption~\ref{assumption:ranks}. If we set $h_n = 1/k_P$ this isn't quite right, unless we break ties weirdly or $k_Q = \Theta(k_P)$.}
\begin{prop}
%Suppose $S \subset [n]$ is chosen uniformly at random. 
1. If $h_n = 1/k_P$ and $k_Q \leq k_P$ then there exists a unique $\Pi \in \{0,1\}^{k_P \times k_Q}$ such that $Z_P \Pi$ contains the $Q$-clustering of all nodes in $[n]$. Let $\Tilde{Z}_Q := Z_P \Pi$. 

2. Let $\hat \Pi$ be as in Algorithm~\ref{alg:q-perfect-clustering} and $U_P, U_Q$ be as in Proposition~\ref{prop:perfect-clustering-assortative}. Then with probability $1 - O(\frac{1}{n_Q})$, $Z_P U_P \hat \Pi = \Tilde{Z}_Q U_Q$. 
\label{prop:perfect-clustering-pi}
\end{prop}
\begin{proof}
Part (1) follows immediately from the SBM structure of $P, Q$ and definition of Definition~\ref{defn:ranks}. 

For Part (2), first notice that by Proposition~\ref{prop:perfect-clustering-assortative}, with probability at least $1 - O(\frac{1}{n_Q})$, Algorithm~\ref{alg:q-perfect-clustering} returns the true clusterings $\hat Z_P = Z_P \in \{0,1\}^{n \times k_P}$ and $\hat Z_Q = Z_Q \in \{0,1\}^{n_Q \times k_Q}$, up to permutation. 

Now, Algorithm~\ref{alg:q-perfect-clustering} simply takes unions of the clusters of $Z_P$ to learn $\hat\Pi$. Therefore, let $V: \RR^{n} \to \RR^{n_Q}$ project onto coordinates in $S$. Then $V \hat Z_P \hat \Pi = \hat Z_Q$. Moreover, by Proposition~\ref{prop:perfect-clustering-assortative}, $\hat Z_P = Z_P U_P$ and $\hat Z_Q = Z_Q U_Q$. Hence $V Z_P U_P \hat \Pi = V \Tilde{Z}_Q U_Q$. To remove dependence on $V$, we need to argue that each $Q$-cluster has a reprensentatve in $S$. 

Let $E$ be the event that at least one $Q$-cluster has no representative in $S$. For a fixed $j \in [k_Q]$, cluster $j$ has no representative in $S$ with probability $\leq \bigg(1 - \frac{n_{min}^{(Q)}}{n_Q} \bigg)^{n_Q}$. A union bound implies: 
\[
\PP[E] \leq k_Q \bigg(1 - \frac{n_{min}^{(Q)}}{n_Q} \bigg)^{n_Q} \leq k_Q \exp(-n_{min}^{(Q)}) \leq O(n_Q^{-1})
\]
The last inequality is because the condition of Theorem~\ref{thrm:chen-cluster} implies that $n_{min}^{(Q)} \geq \Omega(\sqrt{n_Q})$ and $k_Q \leq \frac{n_Q}{n_{min}^{(Q)}}$. 

Finally, we proceed by conditioning on $\lnot E$. Since $\hat Z_P = Z_P U_P$, we know that for all $i \in S$, the unique $j_P \in [k_P]$ such that row $i$, column $j_P$ of $Z_P$ is nonzero contains its true $P$-community up to $U_P$. Similarly since $\hat Z_Q = Z_Q U_Q$, the the unique $j_Q \in [k_Q]$ such that row $i$, column $j_Q$ of $Z_P$ is nonzero contains its true $Q$-community up to $U_Q$. Therefore the nodes in community $j_P$ in $P$ are in community $j_Q$ in $Q$. So, up to permtutations $U_P$ and $U_Q$, we have $\Pi_{j_P, j_Q} = 1$. Since we condition on $\lnot E$, each cluster of $Q$ has at least one representative in $S$, so each columns of $\Pi$ is nonzero. We conclude that $Z_P U_P \hat \Pi = \Tilde{Z}_Q U_Q$ with probability at least $1 - O(n_Q^{-1})$.
\end{proof}

% \blue{Might be cleaner to write things in terms of $Z_P Z_P^T$ rather than $Z_P$, since that is all that is needed to define $\hat Q$. This gets rid of the permutations.}

% \subsubsection{Overall Error of Algorithm~\ref{alg:q-perfect-clustering}}

We are ready to give the overall error of Proposition~\ref{alg:q-perfect-clustering}. 

\begin{prop}
Suppose that $\hat Z_P = Z_P, \hat \Pi = \Pi$ in Algorithm~\ref{alg:q-perfect-clustering}. Then with probability $\geq 1 - O(\frac{1}{n_Q})$, Algorithm~\ref{alg:q-perfect-clustering} returns a $\hat Q \in [0,1]^{n \times n}$ such that:  
\[
\frac{1}{n^2} \norm 
\hat Q - Q
\norm_F^2  
\lesssim
\frac{k_Q^2 \log(n_{min}^{(Q)})}{n_Q^2}
\]
\end{prop}
\begin{proof}
By Proposition~\ref{prop:perfect-clustering-pi}, with probability $\geq 1 - O(\frac{1}{n_Q})$, we have $\hat Z_P = Z_P U_P$, $\hat Z_Q = Z_Q U_Q$, and $\Tilde{Z}_Q U_Q = Z_P U_P \hat \Pi$. We proceed by conditioning on these events. 

Next, let $W_Q \in \RR^{k_Q \times k_Q}$ be the population version of $\hat W_Q$ with $W_{Q;ii} = (\bm{1}^T Z_Q \bm{e}_i)^{-1}$. Then since $\hat Z_Q = Z_Q U_Q$ we have $\hat W_Q = U_Q^T W_Q U_Q$. Hence: 
\begin{align*}
\hat Q &= (Z_P U_P \hat \Pi) (U_Q^T W_Q U_Q^T) (Z_Q U_Q)^T A_Q (Z_Q U_Q) (U_Q^T W_Q U_Q) (Z_P U_P \hat \Pi)^T \\
&= \Tilde{Z}_Q (W_Q Z_Q^T A_Q Z_Q W_Q) \Tilde{Z}_Q^T 
\end{align*}

% \begin{align*}
% W_Q^{-1} \hat Q W_Q^{-1}&= (Z_P U_P \hat \Pi) (Z_Q U_Q)^T A_Q (Z_Q U_Q) (Z_P U_P \hat \Pi)^T \\
% &= \Tilde{Z}_Q U_Q U_Q^T Z_Q^T A_Q Z_Q U_Q U_Q^T \Tilde{Z}_Q \\
% &= \Tilde{Z}_Q (Z_Q^T A_Q Z_Q) \Tilde{Z}_Q^T 
% \end{align*}
Next, let $z_Q: [n] \to [k_Q]$ be the ground truth clustering map given by $\Tilde{Z}_Q \in \{0,1\}^{n \times k_Q}$. Let $B_Q$ be defined analogously to $\hat B_Q$ in Algorithm~\ref{alg:q-perfect-clustering}, but using $W_Q, Z_Q, \EE[A_Q]$ in place of $\hat W_Q, \hat Z_Q, A_Q$. Let $m_i := W_{Q;ii}^{-1}$ be the the number of nodes in $S$ belong to community $i$, and let $n_i$be the the number of nodes in $[n]$ belonging to community $i$ of $Q$. Then the error of Algorithm~\ref{alg:q-perfect-clustering} is then: 
\begin{align*}
\frac{1}{n^2} \norm 
\Tilde{Z}_Q (\hat B_Q  - B_Q) \Tilde{Z}_Q^T
\norm_F^2  
&= \frac{1}{n^2} 
\bigg(
\sum\limits_{i, j \in [k_Q]} n_i n_j
\bigg(\sum\limits_{r \in z_Q^{-1}(\{i\}) \cap S, s \in z_Q^{-1}(\{j\}) \cap S} \frac{B_{Q;ij} - A_{Q;rs}}{m_i m_j}
\bigg)^2
\bigg) \\
&= \frac{1}{n^2} 
\sum\limits_{i, j \in [k_Q]} \frac{n_i n_j}{m_i^2 m_j^2}
\bigg(\sum\limits_{r \in z_Q^{-1}(\{i\}) \cap S, s \in z_Q^{-1}(\{j\}) \cap S} B_{Q;ij} - A_{Q;rs}
\bigg)^2 
\end{align*}
% Where the last step is from Assumption~\ref{assumption:ranks}. 
%Next, let $X_{rs} = (B_{Q;ij} - A_{Q;rs})$ be a single summand.  
% Further, $X_{rs} \in [0,1]$ for all $r, s$.  
Next, fix $i, j \in [k_Q]$ and let $X_{ij} = \sum\limits_{r \in z_Q^{-1}(\{i\}) \cap S, s \in z_Q^{-1}(\{j\}) \cap S} B_{Q;ij} - A_{Q;rs}$ If we condition on the clusterings of $P, Q$ being correct then $\EE[B_{Q;ij} - A_{Q;rs}] = 0$. Therefore by Hoeffding's inequality:
%Moreover, notice all $X_{ij}$ are independent. 
\[
\PP\bigg[
X_{ij}
\geq t^2
\bigg] \leq 2\exp\bigg(\frac{-2t^2}{m_i m_j}\bigg)
\]

Setting $t^2 = 10 \log(m_i m_j) m_i m_j$ implies that with probability at least $1 - k_Q^2 \min\limits_{i} (m_i)^{-20}$, that the overall error is: 
\begin{align*}
\frac{1}{n^2} \norm \hat Q - Q \norm_F^2 &\leq 
\frac{1}{n^2} 
\sum\limits_{i, j \in [k_Q]} \frac{10 \log(m_i m_j) n_i n_j}{m_i m_j}
\end{align*}
Finally, note that there exists constant $c_0$ such that for all $i \in [k_Q]$, $m_i \geq c_0 \sqrt{n_Q}$ and $n_i \geq c_0 \sqrt{n}$, by assumption. Note that each $m_i$ is a random quantity depending on the choice of $S \subset [n]$ such that $\EE[m_i] = \frac{n_Q}{n} n_i$. Hoeffding's inequality and a union bound over all $i \in [k_Q]$ imply that that with probability at least $\geq 1 - O(n_Q^{-8})$ that $m_i \geq \EE[m_i] - 10 \sqrt{\log n_Q} \geq \Omega(\EE[m_i])$. We conclude that: 
\begin{align*}
\frac{1}{n^2} \norm \hat Q - Q \norm_F^2 &\leq 
O\bigg(\frac{1}{n_Q^2} 
\sum\limits_{i, j \in [k_Q]} 10 \log(m_i m_j) 
\bigg) \\
&\leq O(\frac{k_Q^2 \log(n_{min}^{(Q)})}{n_Q})
\end{align*}
\end{proof}
% \blue{should be able to get the error probability.}

% \subsubsection{Overall Error of Algorithm~\ref{alg:q-perfect-clustering}}

% \begin{proof}[Proof of Theorem~\ref{thrm:q-perfect-clustering}]
% Combining the results of ...
% \blue{todo write it out}
% \end{proof}

% !TEX root = ./neurips_2024.tex

\section{Experimental Details}\label{appendix:experiment-details}

In this section we give further details on the experiments of Section~\ref{sec:experiments}. 

{\bf Compute environemnt.} We run all experiments on a personal Linux machine with 378GB of CPU/RAM. The total compute time  across all results in the paper was less than $2$ hours.

{\bf Functions for Heatmap Figure.} For the top row, the source is an $(n,4)$-SBM with $0.8$ on the diagonal and $0.2$ on the off-diagonal of $B \in \RR^{4 \times 4}$. The target is an $(n,2)$-SBM with $0.9$ on the diagonal and $0.1$ on the off-diagonal of $B \in \RR^{2 \times 2}$. 

For the second and third rows, the source function is $Q(x,y) = \frac{1 + \sin(\pi(1 + 3 (x + y - 1)))}{2}$ (modified from \cite{zhang-levina-zhu-2017}). The sources are $P(x,y) = 1 - Q(x,y)$ and $P(x, y) = Q(\phi(x), y)$, where $\phi(x) = 0.5 + \abs{x - 0.5}$ if $x < 0.5$, and $0.5 - \abs{x - 0.5}$ otherwise. 

% and $P(x,y) = Q(\max\{0.5 - \abs{x - 0.5}, 0.5 + \abs{x - 0.5}\}, y)$. We call $Q$ the Period Graphon and $P$ the Inverted Periodic Graphon. 

{\bf Metabolic Networks.} We access metabolic models from \cite{bigg-models} at \url{http://bigg.ucsd.edu}. To construct a reasonable set of shared metabolites across the networks, we take the intersection of the node sets for the following BiGG models: iCHOv1, IJN1463, iMM1415, iPC815, iRC1080, iSDY1059, iSFxv1172, iYL1228, iYS1720, and Recon3D. We obtain a set of $n = 251$ metabolites that are present in all of the listed models.

The resulting networks are undirected, unweighted graphs on $251$ nodes. We construct the matrix $A_P \in \{0,1\}^{n \times n}$ for species $P$ by setting $A_{P;uv} = 1$ if and only if $u$ and $v$ co-occur in a metabolic reaction in the BiGG model for $P$. 

{\bf \textsc{Email-EU}.} We use the ``email-EU-core-temporal'' dataset at \url{https://snap.stanford.edu/data/email-Eu-core-temporal.html}, as introduced in \cite{motifs-2017}. Note that we do not perform any node preprocessing, so we use all $n = 1005$ nodes present in the data, as opposed to \cite{snapnets,motifs-2017} who use only $986$ nodes. 

Data consist of triples $(u, v, t)$ where $u, v$ are anonymized individuals and $t > 0$ is a timestamp. We split the data into $10$ bins based on equally spaced timestamp percentiles. For simplicity we refer to these time periods as consisting of $80$ days each in the main body, but technically there are $803$ days total. The network at time period $\ell$ consists of an unweighted undirected graph with adjacency matrix entry $A_{uv} = 1$ iff $(u, v, t)$ or $(v, t, u)$ occurred in the data for an appropriate timestamp $t$. 

{\bf Hyperparameters.} We do not tune any hyperparameters. For Algorithm~\ref{alg:q-averaging-row-wise} we use the quantile cutoff of $h_n = \sqrt{\frac{\log n_Q}{n_Q}}$ in all experiments.
% How did we construct the metabolic networks - link to BiGG, explain undirected, etc. 

% How did we bin the email networks

% What quantiles did we use everywhere

% Report number of nodes and degrees for all nodes 

% Regarding email: Note that the data as given contain $n_P = 1005$ nodes. Specifically

% Degrees of email data: We report the median of the degrees for each time period. 

% Time Period 0: 6.917412935323383
% Time Period 1: 7.348258706467662
% Time Period 2: 6.667661691542288
% Time Period 3: 7.08457711442786
% Time Period 4: 6.79502487562189
% Time Period 5: 7.099502487562189
% Time Period 6: 7.104477611940299
% Time Period 7: 7.65771144278607
% Time Period 8: 7.262686567164179
% Time Period 9: 7.414925373134328

% For metabolic data: 
% iWFL1372 15.0
% iJN1463 14.0
% iPC815 12.0
%%%%%%%%%%%%%%%%%%%%%%%%%%%%%%%%%%%%%%%%%%%%%%%%%%%%%%%%%%%%

% \newpage
% \section*{NeurIPS Paper Checklist}
% \input{checklist}

\end{document}